\documentclass{article}
\PassOptionsToPackage{numbers, compress}{natbib}
\usepackage[preprint]{neurips_2025}
\usepackage[utf8]{inputenc} %
\usepackage[T1]{fontenc}    %
\usepackage{url}            %
\usepackage{booktabs}       %
\usepackage{amsfonts}       %
\usepackage{nicefrac}       %
\usepackage{microtype}      %
\usepackage{graphicx} \graphicspath{{figures/}}
\usepackage{amsmath,amssymb,mathbbol}
\usepackage{tabularx,colortbl,multirow,array,makecell}
\usepackage{algorithmic}
\usepackage[linesnumbered,ruled,vlined]{algorithm2e}
\usepackage{acronym}
\usepackage{enumitem}
\usepackage[dvipsnames]{xcolor}
\usepackage[pagebackref,breaklinks,colorlinks,citecolor=gray]{hyperref}
\usepackage{xspace}
\usepackage[skip=3pt,font=small]{subcaption}
\usepackage[skip=3pt,font=small]{caption}
\usepackage[capitalise]{cleveref}
\usepackage[misc]{ifsym}
\usepackage{pifont}
\usepackage{wrapfig}

\makeatletter
\DeclareRobustCommand\onedot{\futurelet\@let@token\@onedot}
\def\@onedot{\ifx\@let@token.\else.\null\fi\xspace}
\def\eg{\emph{e.g}\onedot} 

\def\ie{\emph{i.e}\onedot}

\makeatother

\newcommand{\norm}[1]{\left\Vert #1 \right\Vert}

\newcommand{\bx}{\mathbf{x}}
\newcommand{\by}{\mathbf{y}}
\newcommand{\bz}{\mathbf{z}}
\newcommand{\bbR}{\mathbb{R}}
\newcommand{\barbx}{\bar{\bx}}
\newcommand{\hatbx}{\hat{\bx}}

\newcommand{\barbz}{\bar{\bz}}
\newcommand{\hatbz}{\hat{\bz}}
\newcommand{\tilbz}{\tilde{\bz}}

\newcommand{\bzero}{\mathbf{0}}
\newcommand{\cE}{\mathcal{E}}
\newcommand{\cD}{\mathcal{D}}
\newcommand{\hatf}{\hat{f}}

\crefname{algorithm}{Alg.}{Algs.}
\Crefname{algocf}{Algorithm}{Algorithms}
\crefname{section}{Sec.}{Secs.}
\Crefname{section}{Section}{Sections}
\crefname{table}{Tab.}{Tabs.}
\Crefname{table}{Table}{Tables}
\crefname{figure}{Fig.}{Fig.}
\Crefname{figure}{Figure}{Figure}

\makeatletter
\renewcommand{\paragraph}{%
  \@startsection{paragraph}{4}%
  {\z@}{0ex \@plus 0ex \@minus 0ex}{-1em}%
  {\hskip\parindent\normalfont\normalsize\bfseries}%
}
\makeatother

\definecolor{motion-blue}{RGB}{140,192,214}
\definecolor{motion-green}{RGB}{171,209,173}
\definecolor{motion-red}{RGB}{204,20,20}
\definecolor{token-green}{RGB}{186,214,162}
\definecolor{token-yellow}{RGB}{250,225,142}

\acrodef{fsq}[FSQ]{Finite Scalar Quantization}
\acrodef{cvae}[cVAE]{conditional Variational Autoencoder}
\acrodef{rl}[RL]{Reinforcement Learning}
\acrodef{mdp}[MDP]{Markov Decision Process}
\acrodef{dct}[DCT]{discrete cosine transform}
\acrodef{gan}[GAN]{Generative Adversarial Networks}
\acrodef{vae}[VAE]{Variational Autoencoder}
\acrodef{ste}[STE]{straight-through estimator}
\acrodef{hsi}[HSI]{human-scene interaction}

\newcommand{\cmark}{\ding{51}}%
\newcommand{\xmark}{\ding{55}}%
\newcommand{\model}{MSQ\xspace}

\title{Spatial-Temporal Multi-Scale Quantization for Flexible Motion Generation}

\author{%
  Zan Wang$^{1,2*}$, Jingze Zhang$^{2,3*}$, Yixin Chen$^2$, Baoxiong Jia$^{2}$, Wei Liang$^{1,4\dag}$, Siyuan Huang$^{2\dag}$\\
  \small $*$ indicates equal contribution \quad{} $\dag$ indicates corresponding authors\\
  \small $^1$ School of Computer Science \& Technology, Beijing Institute of Technology\\
  \small $^2$ State Key Laboratory of General Artificial Intelligence, BIGAI\\
  \small $^3$ Department of Automation, Tsinghua University\\
  \small $^4$ Yangtze Delta Region Academy of Beijing Institute of Technology, Jiaxing
  \vspace{-16pt}
}

\begin{document}
\maketitle

\begin{figure}[ht!]
    \centering
    \includegraphics[width=\linewidth]{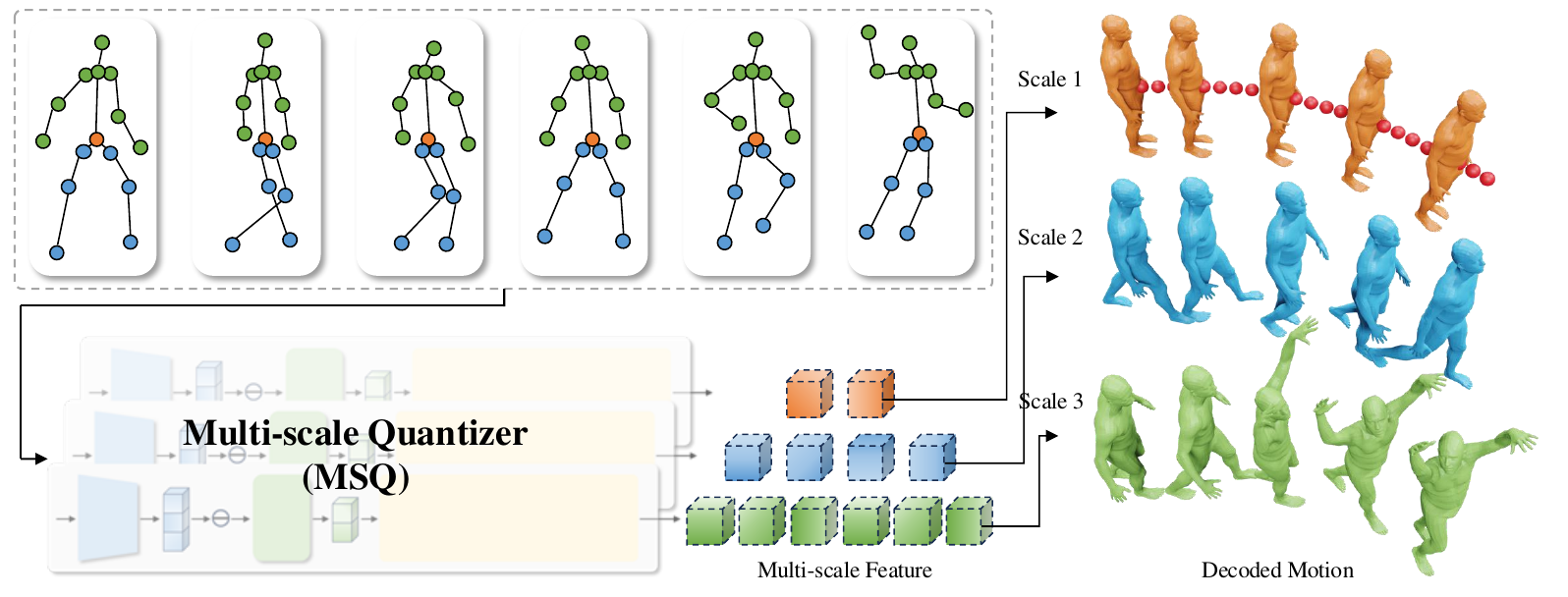}
    \caption{We propose \model, a novel quantization method that compresses human motion into a multi-scale discrete token space across temporal and spatial dimensions, facilitating the effectiveness and flexibility in diverse motion modeling tasks. Each scale captures different granularities of motion features, resulting in distinct motion sequences upon decoding. Our project website is available at \href{https://sites.google.com/view/msquantizer}{https://sites.google.com/view/msquantizer}.}
    \label{fig:teaser}
\end{figure}

\begin{abstract}
Despite significant advancements in human motion generation, current motion representations---typically formulated as discrete frame sequences---still face two critical limitations: (i) they fail to capture motion from a multi-scale perspective, limiting the capability in complex patterns modeling; (ii) they lack compositional flexibility, which is crucial for model's generalization in diverse generation tasks.
To address these challenges, we introduce \model, a novel quantization method that \textbf{compresses the motion sequence into multi-scale discrete tokens across spatial and temporal dimensions.} \model employs distinct encoders to capture body parts at varying spatial granularities and temporally interpolates the encoded features into multiple scales before quantizing them into discrete tokens. Building on this representation, we establish a generative mask modeling model to effectively support motion editing, motion control, and conditional motion generation.
Through quantitative and qualitative analysis, we show that our quantization method enables the seamless composition of motion tokens without requiring specialized design or re-training. Furthermore, extensive evaluations demonstrate that our approach outperforms existing baseline methods on various benchmarks.
\end{abstract}

\section{Introduction}
\label{sec:intro}

Human motion generation~\citep{petrovich2021action,guo2022generating,petrovich2022temos,tevet2023human,zhang2023generating,li2023object,jiang2024scaling,guo2024momask} has progressed significantly in the past few years, driving various applications in fields like virtual reality, video games, and film production. Recent advancements have also pivoted towards practical objectives such as motion editing~\citep{zhang2023finemogen,athanasiou2024motionfix,goel2024iterative} and motion control~\citep{karunratanakul2023guided,rempe2023trace,xie2024omnicontrol}, empowering creators to customize motion outputs to their specific requirements. Supporting these diverse tasks requires a comprehensive and flexible motion representation---a long-standing goal pursued by the community for years.

Current methods~\citep{cao2020long,guo2022generating,li2023object,wang2022humanise,huang2023diffusion,diller2024cg} predominantly rely on skeleton-based representations, where human motion is defined as a sequence of single-frame body poses, modeled by joint positions or statistical parametric models like SMPL~\citep{loper2015smpl}. 
Some approaches~\citep{zhang2023generating,lu2023humantomato,chen2023executing,dai2024motionlcm} further compress these data into continuous or discrete latent spaces via \ac{vae}~\citep{kingma2013auto} or VQ-\ac{vae}~\citep{van2017neural} to facilitate the learning process of generative models. Despite progress, these representations still encounter two fundamental limitations, failing to capture the complex dynamics necessary for effective and flexible motion modeling.

First, current motion representations lack multi-scale modeling across both temporal and spatial dimensions.
Multi-scale modeling, interpreting data at multiple levels, has proven essential in capturing complex patterns for image modeling~\citep{lin2017feature,karras2017progressive,tian2024visual}.
In the context of human motion, multi-scale modeling enables models to capture high-level semantic patterns and fine-grained dynamic details over time. Spatially, it distinguishes coarse motion trajectories from individual joint movements, offering a more nuanced understanding of motion structure.

Second, existing methods struggle to handle compound motions due to insufficient compositional capacity. A representation with a robust compositional structure would enable generative models to generalize to unseen motion compositions, significantly enhancing the flexibility and precision of diverse fine-grained motion editing and part-based motion control tasks. Though recent approaches~\citep{pi2023hierarchical,zhong2023attt2m,wan2024tlcontrol} decompose the whole body into parts for feature learning in the latent space, they remain intractable when directly composing different motion components.

To address these issues, \textbf{\textit{we introduce a novel motion quantization method, denoted as \model, which compresses motion sequences into multi-scale discrete tokens across spatial and temporal dimensions,}} as illustrated in \cref{fig:teaser}. Once encoding the motion into discrete tokens, we model the motion distribution through a generative masked Transformer~\citep{chang2022maskgit,guo2024momask,pinyoanuntapong2024mmm}, enabling diverse applications in motion editing, motion control, and conditional motion generation. Specifically, \model uses separate CNN-based encoders to capture decomposed body parts at varying spatial granularities, followed by corresponding interpolation operations at different temporal scales. This technique depicts the coarse motion features like the pelvis trajectories at a higher level with fewer tokens, while finer features like limb dynamics are the opposite. Rather than relying on conventional vector quantization~\citep{zhang2023generating,guo2024momask}, we adopt \ac{fsq}~\citep{mentzer2024finite} to quantize latent features into discrete token indices, mitigating the code index collapse during training. For reconstruction, \model merges the dequantized codes and projects them back into the original motion space via a single decoder network.

Through quantitative and qualitative analysis, we showcase how the feature at each scale contributes to the motion representation and how \model supports the seamless composition of motion tokens, which enables versatile motion editing and control tasks without requiring model modifications or fine-tuning. We also conduct extensive experiments across diverse benchmarks to demonstrate our method's superiority. Our method outperforms the state-of-the-art text-based motion editing methods on the MotionFix~\citep{athanasiou2024motionfix}. For conditional motion generation, our method achieves superior performance on the HumanML3D~\citep{guo2022generating} and significantly enhances language-guided \ac{hsi} generation on the HUMANISE~\citep{wang2022humanise}.

Our contributions are summarized as follows: (1) We propose a novel motion quantization method that compresses motion sequences into a multi-scale discrete token space, integrating generative masked modeling to support diverse motion modeling tasks. (2) We analyze the impact of each scale within the proposed multi-scale representation and showcase its inherent capacity for seamless motion composition, which enables flexible motion editing and control. (3) Extensive evaluations demonstrate our method's superiority over existing methods on various benchmarks, including MotionFix, HumanML3D, and HUMANISE.

\section{Related Work}
\label{sec:related work}

\paragraph{Human Motion Generation}
Early work on human motion generation primarily focused on motion forecasting given historical motion sequences~\citep{mao2020history,yuan2020dlow,cao2020long}. Leveraging annotated motion data, several studies~\cite{ahuja2019language2pose,guo2020action2motion,petrovich2021action,petrovich2022temos,guo2022generating,zhang2023remodiffuse,tevet2023human,zhang2023generating,chen2023executing,pinyoanuntapong2024mmm} introduced methods to generate diverse and realistic human motions conditioned on semantic cues like action labels and text descriptions.
Recently, efforts have extended to other conditional modalities, such as audio~\citep{zhang2024bidirectional,chhatre2024emotional}, object ~\citep{zhang2022couch,li2023object,jiang2023full,diller2024cg}, 3D scenes~\citep{araujo2023circle,yi2024generating,jiang2024scaling,wang2024move}, and interactive movements~\citep{liang2024intergen,xu2024inter}, further booming motion generation research. These approaches often model motion probability distributions using generative models like diffusion models~\cite{tevet2023human,chen2023executing,huang2023diffusion}. Another direction~\citep{peng2021amp,xiao2023unified,cui2024anyskill} formulates motion generation as a \ac{mdp}, applying \ac{rl} to learn policies for avatar movement control. Besides conditional motion generation, motion editing has been a long-standing focus, including style transfer~\citep{aberman2020unpaired,jang2022motion,song2024arbitrary} and recent text-guided editing of specific body parts~\citep{zhang2023finemogen,goel2024iterative,athanasiou2024motionfix}. With diffusion models, some methods~\citep{karunratanakul2023guided,rempe2023trace,xie2024omnicontrol} incorporate spatial guidance (\eg, trajectories and joint positions) into the denoising process to enable control over motion generation.
We introduce a multi-scale motion representation to support diverse motion generation tasks with enhanced effectiveness and flexibility. Besides, the inherent composition capability allows the seamless composition of motion tokens without requiring model modifications or re-training.

\paragraph{Motion Representation Modeling}
Most works~\citep{yuan2020dlow,cao2020long,guo2022generating,li2023object} represent motion as a sequence of joint positions or rotations, often called a skeleton-based representation. Some methods~\citep{mao2020history,zhang2021we} transform this spatial-temporal data into the frequency domain using \ac{dct}~\citep{ahmed1974discrete} to improve learning efficiency. Other methods~\citep{petrovich2021action,wang2022humanise,huang2023diffusion,diller2024cg} leverage statistical parametric models like SMPL(-X)~\citep{loper2015smpl,pavlakos2019expressive}, which capture body translation, orientation, pose, and shape. 
\citet{rempe2021humor,zhang2021learning} learn motion prior from existing datasets, often serving as regularization terms to further enhance motion modeling.
To enable auto-regressive modeling, \citet{zhang2023generating,lu2023humantomato} quantize motion frames into discrete tokens using VQ-\acs{vae}; in contrast, \citet{chen2023executing,dai2024motionlcm} employ a \acs{vae} to compress motion into a latent space, followed by a latent diffusion model for generation. Recent methods \citep{pi2023hierarchical,zhong2023attt2m,wan2024tlcontrol,sun2024lgtm} decompose the body into separate parts to achieve expressive features for motion generation and control.
\citet{pi2023hierarchical,wan2024tlcontrol} quantize per-frame body parts into discrete tokens; however, they neglect the multi-scale properties inherent in the motion data, reducing the modeling capability and flexibility.
Unlike these works, our approach models motion in a discrete latent space with multi-scale features across spatial and temporal dimensions, possessing improved effectiveness and flexibility for various motion modeling tasks.

\paragraph{Generative Modeling}
Over the past decades, the computer vision community has explored various generative models for image generation, starting with \ac{gan}~\citep{goodfellow2014generative,karras2019style} and progressing to diffusion models~\citep{ho2020denoising,dhariwal2021diffusion,rombach2022high}, which now dominate the field. Recently, auto-regressive models~\citep{esser2021taming,yu2022scaling,tian2024visual} and generative masked modeling~\citep{devlin2018bert,chang2022maskgit,chang2023muse,li2023mage} have also gained significant attention for their high-quality generation capabilities. The motion generation field has followed a similar developmental trajectory in many ways.
Early methods~\citep{petrovich2021action,petrovich2022temos,wang2022humanise} primarily rely on \ac{cvae} for generating motion conditioned on inputs like text descriptions, especially when the dataset scale is limited. MDM~\citep{tevet2023human} is the first to introduce diffusion models into human motion generation, inspiring a wave of follow-up research~\citep{chen2023executing,huang2023diffusion,dai2024motionlcm}. Some approaches~\citep{zhang2023generating,lu2023humantomato,jiang2023motiongpt} represent motion frames as discrete tokens and use decoder-only transformers to generate motion auto-regressively. More recently, generative masked modeling~\citep{guo2024momask,pinyoanuntapong2024mmm} has demonstrated notable improvements over diffusion and auto-regressive models, paving the way for new advancements in the field.
This work integrates the proposed multi-scale quantization with generative masked modeling, 
achieving improved performance on diverse benchmarks~\citep{athanasiou2024motionfix,guo2022generating,wang2022humanise}.

\section{Preliminaries}
\label{sec:preliminary}

\paragraph{\acf{fsq}} \ac{fsq}~\citep{mentzer2024finite}, as a drop-in replacement for vector quantization, avoids issues like codebook collapse and the need for complex mechanisms like commitment losses~\citep{van2017neural} and code reinitialization~\citep{dhariwal2020jukebox}. Given a vector $z \in \mathbb{R}^d$, \ac{fsq} maps each entry $z_i$ to a discrete space with $L$ values through a bounding function $f:z\mapsto\lfloor \nicefrac{L}{2} \rfloor\texttt{tanh}(z)$ followed by a rounding operation, finally yielding a quantized vector $\hat{z}=\texttt{round}(f(z))$. Assuming the $i$-th channel is mapped to $L_i$ values, \ac{fsq} ultimately results in an implied codebook $\mathcal{C}$ with size $|\mathcal{C}|=\prod_{i=1}^dL_i$. Additionally, \ac{fsq} employ \ac{ste}~\citep{bengio2013estimating} to propagate gradients through the $\texttt{round}$ operation by applying the stop-gradient mechanism, \ie, $\texttt{round\_ste}: z \mapsto z+\texttt{sg}(\texttt{round}(f(z)) - z)$. For more details, we recommend referring to the original paper.

\paragraph{Masked Transformer}
Generative masked modeling~\citep{devlin2018bert,chang2022maskgit,chang2023muse,li2023mage} is a self-supervised learning technique where a model learns to predict a set of random masked tokens based on the known tokens. Recent approaches~\citep{chang2022maskgit,guo2024momask,pinyoanuntapong2024mmm} have employed this learning paradigm to train bidirectional masked transformers for content generation in an auto-regressive manner.
Specifically, during training, a random subset of content tokens is sampled and replaced with the special $\texttt{[MASK]}$ token, with the masking ratio drawn from a pre-defined scheduling function $\gamma(r) \in (0,1]$, where $r$ is uniformly sampled. The masked transformer then learns to reconstruct the masked tokens from the observed ones.
During sampling, the model begins with a fully masked token sequence and generates all tokens auto-regressively in two steps: (1) It generates a complete token sequence using the masked transformer. (2) It re-masks $\lceil \gamma(\nicefrac{k}{K}) \cdot n \rceil$ tokens with the lowest confidence while maintaining the other tokens unchanged for subsequent iterations. $n$ is the number of content tokens, and $k$ is the current iteration step. This sampling process continues for $K$ iterations.

\section{Method}
\label{sec:method}

\begin{figure*}[t!]
    \centering
    \includegraphics[width=\linewidth]{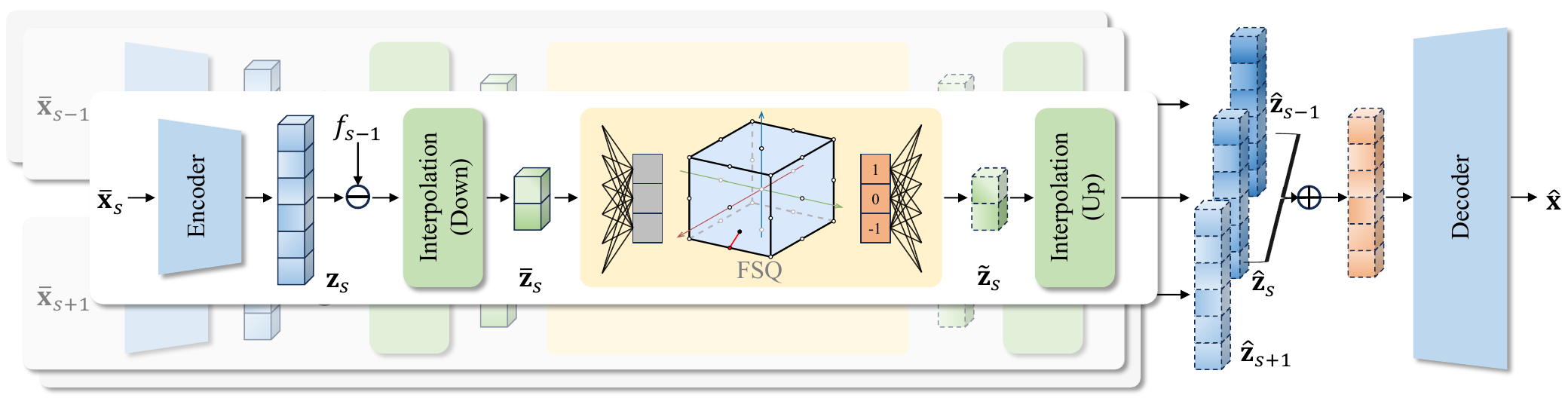}
    \caption{\textbf{Overview of \model.} \model separately compresses motion features at various spatial and temporal granularities, producing a multi-scale discrete token representation. These dequantized features are merged and passed to a decoder for motion reconstruction.}
    \label{fig:quantizer}
\end{figure*}

We propose a novel motion quantization method, \ie, \model, which encodes human motion into multi-scale discrete tokens across spatial and temporal dimensions. \cref{fig:quantizer} illustrates the design of \model. Building on this multi-scale representation, we develop a masked transformer to support diverse motion modeling tasks. Below, we provide a detailed description of each component.

\subsection{Spatial-Temporal Multi-Scale Quantization}

Given a human motion sequence $\bx \in \bbR^{N \times D}$, where $N$ is the number of motion frames and $D$ is the dimensionality of each frame's pose representation, \model aims to compress $\bx$ into a discrete token sequence $\by$ with a multi-scale structure.

\paragraph{Encoding} To achieve a multi-scale representation, \model first follows prior works~\citep{pi2023hierarchical,wan2024tlcontrol} to decompose $\bx$ into $S$ sub-features, denoted as  $\left\{\barbx_s\right\}_{s=1}^{S}$. Each sub-feature $\barbx_s \in \bbR^{N \times D_s}$ captures motion features related to specific body parts. Here, $D_s \leq D_{s+1}$ allows each feature $\barbx_s$ to represent an increasingly finer granularity of motion, with coarser scales capturing coarse motion trends and finer scales retaining detailed motion nuances. For instance, the coarsest scale feature $\barbx_1$ captures the motion trajectory through the pelvis joint, providing a rough depiction of motion; full-body features are retained at the finest scale (\ie, $s=S$), ensuring precise representation of detailed motion.
We then encode $\barbx_s$ into a latent representation $\bz_s \in \mathbb{R}^{n \times d}$ using a unique convolutional encoder per scale, following ~\citet{pi2023hierarchical,wan2024tlcontrol}. Here, $\nicefrac{n}{N}$ is the down-sampling factor, and $d$ represents the dimensionality of the latent space.

Next, we interpolate $\bz_s$ along the temporal dimension to produce vector $\barbz_s = \texttt{Inter}(\bz_s)$, where $\barbz_s \in \mathbb{R}^{n_s \times d}$ and $\texttt{Inter}(\cdot)$ is the interpolation operation, yielding a sequence of multi-scale latent feature $\left\{\barbz_s\right\}_{s=1}^{S}$. By setting $n_s \leq n_{s+1}$, we ensure that fewer tokens represent high-level motion information while progressively more tokens for low-level details, enhancing the expressiveness of the representation.

\paragraph{Quantization}
We apply \ac{fsq}~\cite{mentzer2024finite} to quantize each latent vector $\barbz_s$, yielding a quantized vector $\tilbz_s = \texttt{FSQ}(\barbz_s)$, where $\tilbz_s \in \bbR^{n_s \times d}$. Each quantized vector $\tilbz_s$ corresponds to a discrete index sequence $\by_s \in \left\{1,\cdots,|\mathcal{C}|\right\}^{n_s}$. We concatenate index sequences of all scales (\ie, $\left\{\by_s\right\}_{s=1}^{S}$) to form the final index sequence $\by$.

\paragraph{Decoding}
To reconstruct the original motion sequence, we begin by interpolating each quantized code $\tilbz_s$ back to $\hatbz_s \in \bbR^{n \times d}$. Next, we merge the codes $\left\{\hatbz_s\right\}_{s=1}^{S}$ and forward the result through a decoder implemented with de-convolution layers to produce the reconstructed motion $\hatbx$. 
Inspired by \citet{lee2022autoregressive}, we improve learning efficiency by interpolating and quantizing the residual features instead of directly quantizing $\barbz_s$ in the implementation. Specifically, we apply $\tilbz_s = \texttt{FSQ}(\texttt{Inter}(\bz_s - f_{s-1}))$ and define $f_s=\sum_{i=1}^{s} \hatbz_i$.
This approach allows us to aggregate the dequantized residual features $\hatbz_s$ across all scales through simple addition. We pass the final aggregated result $f_S$ to the decoder for motion reconstruction.

\paragraph{Training Objective}

We train \model end-to-end using the following loss function:
\begin{equation}
    \mathcal{L} = \norm{\bx - \hatbx} + \alpha \sum_{s=1}^{S} \left(\norm{\bz_s - \texttt{sg}(f_s)} + \norm{\texttt{sg}(\bz_s) - f_s}\right).
\end{equation}
Here, $\texttt{sg}(\cdot)$ denotes the stop gradient, and $\alpha=0.1$ specifies the loss weight. While FSQ~\cite{mentzer2024finite} removes the need for VQ objective and commitment loss terms, we retain them to accelerate convergence.

\subsection{Mask Modeling for Diverse Tasks}

\begin{wrapfigure}[17]{r}{0.54\linewidth}
    \vspace{-2pt}
    \centering
    \includegraphics[width=\linewidth]{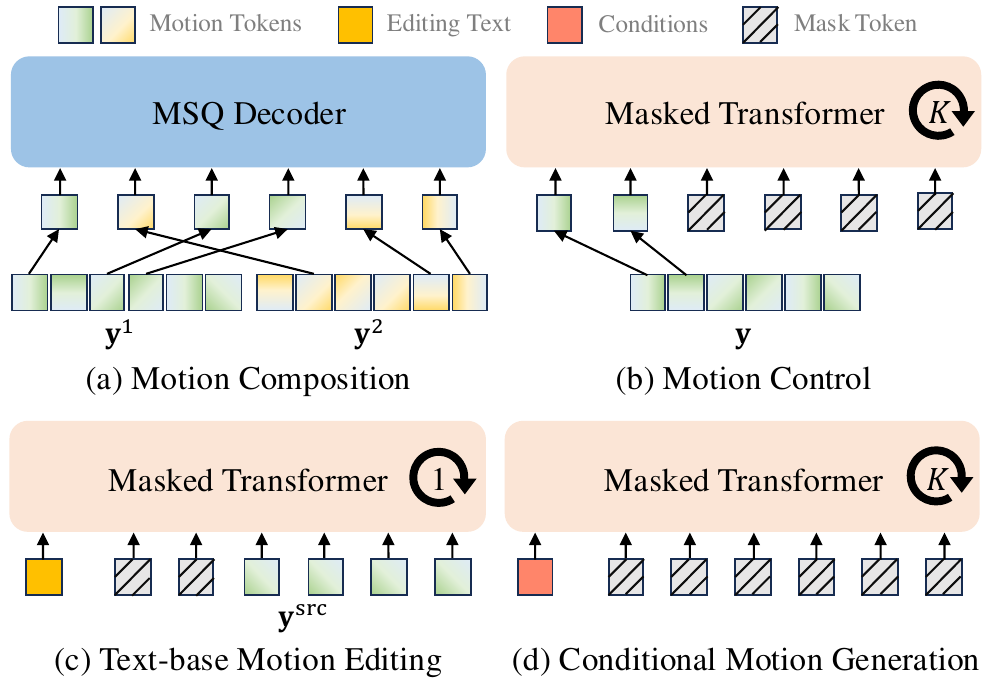}
    \caption{\textbf{Task-specific adaptations of our model for diverse tasks.} The light \textcolor{token-green}{green} and \textcolor{token-yellow}{yellow} colors indicate that the tokens are derived from two different motions.}
    \label{fig:transformer}
\end{wrapfigure}

Building on \model, we incorporate generative masked modeling to support diverse motion generation tasks by training a bidirectional masked transformer, as shown in \cref{fig:transformer}. Specifically, after quantizing motions from the dataset into discrete tokens, the model learns to generate motion tokens with or without other conditions by modeling the tokens' distribution in an auto-regressive manner. The training objective is to minimize the negative log-likelihood of the target token predictions.
Below, we describe the model’s task-specific adaptations in detail.

\paragraph{Motion Composition}
\model inherently supports flexible motion composition by directly combining tokens from different motion sequences, as shown in \cref{fig:transformer}\textcolor{red}{a}. For instance, given two motion sequences, $\bx^1$ and $\bx^2$, we quantize them into token sets $\left\{\by_s^1\right\}_{s}^{S}$ and $\left\{\by_s^2\right\}_{s}^{S}$. To create a composite sequence, we concatenate tokens as $\left\{\left[\by^1_{s,1:i}; \by^2_{s,i+1:n_s}\right]\right\}_{s=1}^{S}$, enabling motion blending along the temporal dimension, as the token order corresponds to temporal progression. We then decode the composed tokens to a motion sequence that resembles $\bx^1$ in the initial frames and transitions to resemble $\bx^2$ in the subsequent frames. For spatial composition, we similarly combine tokens across different scales, $\left[\by^1_{1:s}; \by^2_{s+1:S}\right]$. This composition results in a motion that captures $\bx^1$ partial parts' dynamics, primarily represented by scales $1$ to $s$ with residual modeling, while the remaining parts resemble $\bx^2$.

\paragraph{Motion Control}
Once \model and the masked transformer are trained, our method enables motion control generation by providing a pelvis trajectory without requiring model modification or re-training. Specifically, we directly quantize the concatenation of pelvis trajectory and zero paddings into discrete tokens. By taking the first scale tokens corresponding to the pelvis and other scales' masked tokens, the pre-trained masked transformer recovers the masked tokens, resulting in a motion sequence that aligns well with the specified trajectory, as illustrated in \cref{fig:transformer}\textcolor{red}{b}.

\paragraph{Text-based Motion Editing}
For text-based motion editing (\ie, the MotionFix~\citep{athanasiou2024motionfix} setting), we use \model to quantize both source and target motions into discrete tokens.
Given the editing text and source motion tokens $\by^{\text{src}}$ as input, we train the transformer to predict the target tokens $\by^{\text{tar}}$ in a \textit{single} iteration. As shown in \cref{fig:transformer}\textcolor{red}{c}, we can optionally mask the editing part of the source tokens.

\paragraph{Conditional Motion Generation}
For conditional motion generation in \cref{fig:transformer}\textcolor{red}{d}, we directly concatenate the conditional token embeddings (\eg, derived from CLIP~\citep{radford2021learning} for text-to-motion task) with the fully masked motion tokens. This concatenation is then passed through the masked transformer to predict the ground truth tokens in $K$ iterations.

\subsection{Implementation}

We train both \model and the masked transformer using a \texttt{CosineAnnealingWarmup} scheduler~\citep{scheduler} with a maximum learning rate of $10^{-4}$ and a minimum of $10^{-5}$, alongside the \texttt{Adam} optimizer. \model is trained on 2 Nvidia RTX 3090 GPUs with a batch size of 128 per GPU, while the masked transformer is trained on 4 Nvidia RTX 3090 GPUs with a batch size of 16 per GPU. 
To schedule the training of the masked transformer, we follow prior works~\citep{chang2022maskgit,guo2024momask} to adopt a \texttt{cosine} function $\gamma(\cdot)$. By default, we fix $S=6$ in \model and $K=5$ during the sampling.

\section{Experiments}
\label{sec:exp}

This section begins with a comprehensive analysis of the proposed multi-scale motion representation, followed by extensive quantitative and qualitative evaluations across a range of benchmarks.

\subsection{Analysis of Multi-Scale Representation}

To analyze the capability of \model, we adopt the HumanML3D~\citep{guo2022generating} as the training data source. Details about the dataset and evaluation metrics are illustrated in \cref{sec:eval_h3d}.

\paragraph{Impact of Per Scale on Reconstruction}
We analyze the impact of each scale in our multi-scale representation through two ablative validations: (1) evaluating the contribution of the first $s$ scales to motion reconstruction by merging and decoding tokens only from these scales, denoted as ``\model($1$, $s$)''; (2) assessing the effect of each individual scale on the final reconstruction by removing the tokens at that scale, denoted as ``w/o \model($s$)''.

\begin{table}[t!]
    \centering
    \small
    \caption{\textbf{The scale impact on reconstruction quality.}}
    \label{tab:ablate_scale}
        \begin{tabular}{ccccc}%
            \toprule
            Model & \makecell{R-Precision\\(Top 1)} $\uparrow$ & FID $\downarrow$ & MM Dist. $\downarrow$ & MPJPE$\downarrow$ \\
            \midrule
            Real              & $0.511^{\pm.003}$ & $0.002^{\pm.000}$ & $2.974^{\pm.008}$ & $0.0^{\pm.000}$    \\
            \midrule
            \model($1$,$1$)   & $0.279^{\pm.003}$ & $5.781^{\pm.031}$ & $4.589^{\pm.011}$ & $108.63^{\pm.083}$ \\
            \model($1$,$2$)   & $0.323^{\pm.005}$ & $5.554^{\pm.017}$ & $4.413^{\pm.017}$ & $87.88^{\pm.130}$  \\
            \model($1$,$3$)   & $0.378^{\pm.000}$ & $5.285^{\pm.016}$ & $4.021^{\pm.014}$ & $68.10^{\pm.055}$  \\
            \model($1$,$4$)   & $0.494^{\pm.000}$ & $0.158^{\pm.002}$ & $3.042^{\pm.036}$ & $44.60^{\pm.034}$  \\
            \model($1$,$5$)   & $0.505^{\pm.000}$ & $0.077^{\pm.001}$ & $2.966^{\pm.011}$ & $34.41^{\pm.036}$  \\
            \midrule
            w/o \model($1$)   & $0.446^{\pm.000}$ & $1.060^{\pm.005}$ & $3.286^{\pm.012}$ & $68.98^{\pm.138}$ \\
            w/o \model($2$)   & $0.483^{\pm.000}$ & $0.296^{\pm.003}$ & $3.051^{\pm.012}$ & $50.67^{\pm.078}$ \\
            w/o \model($3$)   & $0.472^{\pm.004}$ & $0.509^{\pm.003}$ & $3.209^{\pm.010}$ & $49.69^{\pm.042}$ \\
            w/o \model($4$)   & $0.478^{\pm.003}$ & $0.347^{\pm.002}$ & $3.053^{\pm.010}$ & $43.10^{\pm.043}$ \\
            w/o \model($5$)   & $0.501^{\pm.004}$ & $0.093^{\pm.001}$ & $2.984^{\pm.011}$ & $36.63^{\pm.040}$ \\
            w/o \model($6$)   & $0.505^{\pm.000}$ & $0.077^{\pm.001}$ & $2.966^{\pm.011}$ & $34.41^{\pm.036}$ \\
            \midrule
            Full Model        & $\mathbf{0.511^{\pm.000}}$ & $\mathbf{0.037^{\pm.000}}$ & $\mathbf{2.934^{\pm.012}}$ & $\mathbf{23.67^{\pm.050}}$ \\
            \bottomrule
        \end{tabular}%
\end{table}

The quantitative results are presented in \cref{tab:ablate_scale}. The results show that reconstruction performance improves as the number of scales used for decoding increases. Notably, metrics like \textit{FID} show a sudden increase when the fourth scale works. \textbf{We speculate that this scale incorporates features of the arms, whose dynamic richness significantly contributes to these metrics.} Ablation studies for individual scales also reveal that the first scale, which primarily encodes pelvis information, is the most critical, with the impact of subsequent scales gradually diminishing.

\begin{figure*}[t!]
    \centering
    \includegraphics[width=\linewidth]{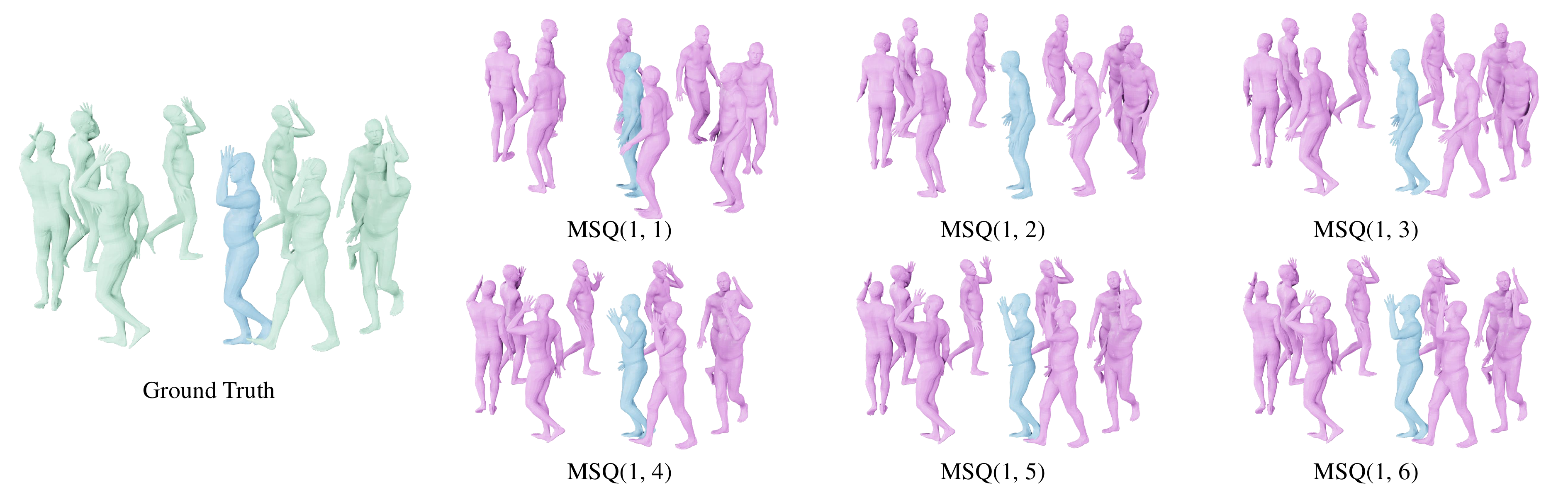}
    \caption{\textbf{Multi-scale visualization.} We present the reconstruction results of model ``MSQ($1$, $s$)''. The lower scales capture the coarse motion trends, while higher scales depict the motion in a finer granularity. The pose in \textcolor{motion-blue}{blue} denotes the first frame of each motion sequence.}
    \label{fig:multi-scale}
\end{figure*}

We visualize the reconstruction results of ``\model ($1$, $s$)'' in \cref{fig:multi-scale}. From the visualizations, we observe that the lower scales capture the coarse motion trends, while higher scales provide a finer-grained depiction of the motion. For instance, the reconstruction result of ``MSQ($1$, $1$)'' reflects only the rough trajectory of the motion, neglecting the finer limb dynamics. In contrast, the result of ``MSQ($1$, $6$)'' closely matches the ground truth. These findings highlight the effectiveness of the multi-scale structure in our proposed representation, which potentially facilitates the flexibility in handling diverse motion modeling tasks.

\begin{figure*}[t!]
    \centering
    \begin{subfigure}{0.8\linewidth}
        \includegraphics[width=\linewidth]{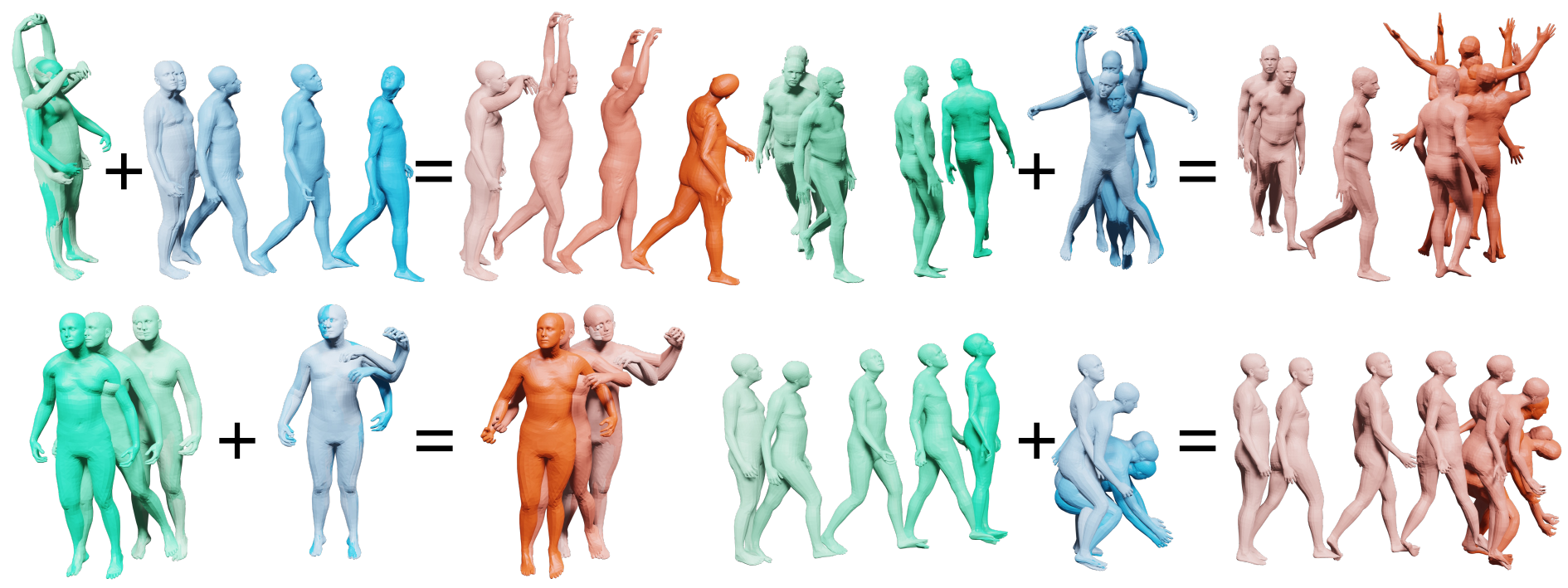}
        \caption{Motion Composition}
        \label{fig:compose}
    \end{subfigure}%
    \begin{subfigure}{0.2\linewidth}
        \includegraphics[width=\linewidth]{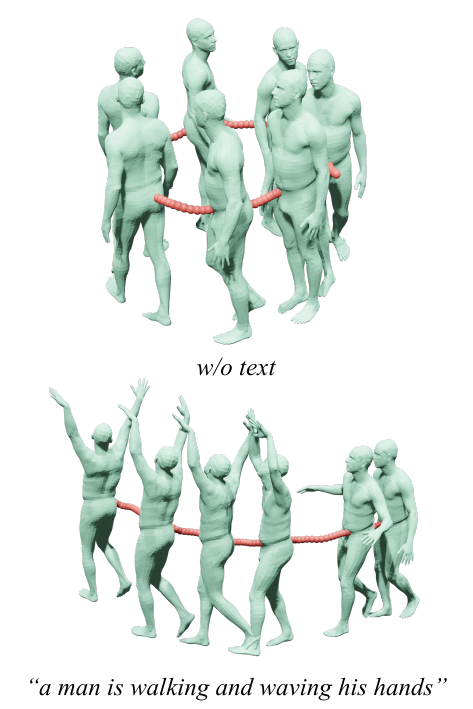}
        \caption{Motion Control}
        \label{fig:control}
    \end{subfigure}%
    \caption{\textbf{Motion composition and control.} (a) Given two motion sequences, shown in \textcolor{motion-green}{green} and \textcolor{motion-blue}{blue}, \model's inherent capability enables the seamless composition of corresponding tokens for spatial (left two cases) and temporal (right two cases) motion composition. (b) Our method can also generate controllable motions aligning well with the given pelvis trajectory (\textcolor{red}{red} balls), with or without text conditions.}
    \label{fig:composition}
\end{figure*}

\paragraph{Motion Composition and Control}
We qualitatively demonstrate the composition capability of our representation in \cref{fig:composition}.
Our method enables the direct composition of two motion sequences along temporal or spatial dimensions, yielding plausible outcomes that capture partial dynamics from both motions, as shown in \cref{fig:compose}.
Additionally, this capability allows the decomposition of body parts at specific scales, such as capturing pelvis information in the first scale. We quantize the pelvis trajectory into discrete tokens and use them to control motion generation via a pre-trained masked transformer. \cref{fig:control} presents the generated results that closely align with the guided pelvis trajectory, regardless of whether using text conditions.

\begin{table*}[t!]
    \centering
    \small
    \caption{\textbf{Quantitative results on the MotionFix~\citep{athanasiou2024motionfix}.} ``Real'' denotes results computed from ground truth. \textbf{Bold} indicates the best result.}
    \label{tab:motionfix}
    \resizebox{\linewidth}{!}{%
        \begin{tabular}{ccccccccccc}
            \toprule
            \multirow{2.4}{*}{Model} & \multirow{2.4}{*}{Data} & \multirow{2.4}{*}{Source Input} & \multicolumn{4}{c}{generated-to-target retrieval} & \multicolumn{4}{c}{generated-to-source retrieval} \\
            \cmidrule(lr){4-7} \cmidrule(lr){8-11} & & & R@1$\uparrow$ & R@2$\uparrow$ & R@3$\uparrow$ & AvgR$\downarrow$ & R@1$\uparrow$ & R@2$\uparrow$ & R@3$\uparrow$ & AvgR$\downarrow$ \\
            \midrule
            Real & -  & -  & $100.0$ & $100.0$ & $100.0$ & $1.00$ & $74.01$ & $84.52$ & $89.91$ & $2.03$ \\
            \midrule
            MDM                                  & HumanML3D  & \xmark & $4.03$  & $7.56$  & $10.48$ & $15.55$ & $2.62$  & $6.15$  & $9.38$  & $15.88$ \\
            MDM\textsubscript{S}                 & HumanML3D  & \cmark & $3.63$  & $7.06$  & $10.08$ & $15.64$ & $2.62$  & $6.25$  & $9.78$  & $15.84$ \\
            MDM-BP                               & HumanML3D  & \xmark & $39.10$ & $50.09$ & $54.84$ & $6.46$  & $61.28$ & $69.55$ & $73.99$ & $4.21$  \\
            MDM-BP\textsubscript{S}              & HumanML3D  & \cmark & $38.10$ & $48.99$ & $54.84$ & $6.47$  & $60.28$ & $69.46$ & $73.89$ & $4.23$  \\
            \midrule
            MDM~\citep{tevet2023human}           & MotionFix & \cmark & $18.54$          & $29.17$          & $36.04$          & $11.91$         & $15.01$                   & $23.75$          & $31.04$          & $11.96$        \\
            TMED~\citep{athanasiou2024motionfix} & MotionFix & \cmark & $62.90$          & $76.51$          & $83.06$          & $2.71$          & $71.77$                   & $84.07$          & $89.52$          & $1.96$          \\
            Ours                                 & MotionFix & \cmark & $\mathbf{68.18}$ & $\mathbf{82.62}$ & $\mathbf{87.64}$ & $\mathbf{2.30}$ & $\mathbf{74.50}$          & $\mathbf{87.22}$ & $\mathbf{91.17}$ & $\mathbf{1.92}$ \\
            \bottomrule
        \end{tabular}
    }%
\end{table*}

\begin{figure*}[t!]
    \centering
    \includegraphics[width=\linewidth]{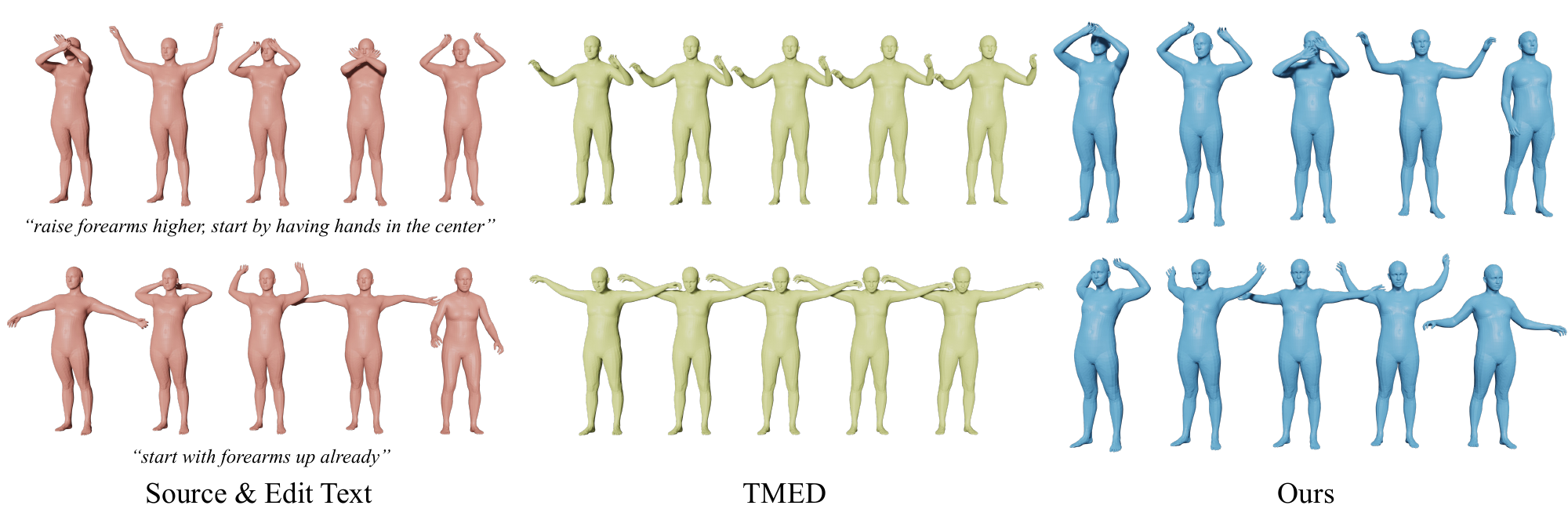}
    \caption{\textbf{Qualitative results on MotionFix dataset.} Given editing texts and source motions, our method successfully generates plausible target motions, while the state-of-the-art method TMED~\citep{athanasiou2024motionfix} fails to effectively align the edit parts with descriptions.}
    \label{fig:motionfix}
\end{figure*}

\subsection{Evaluation on MotionFix}

\paragraph{Setting, Metrics, and Baselines}
MotionFix~\citep{athanasiou2024motionfix} is a recently established text-based motion editing benchmark, requiring generation models to produce a plausible target human motion given a source motion and edited text prompt.
It includes $6,730$ source-target motion pairs with text annotations for training and evaluation.
We use MotionFix’s original raw pose representations and dataset splits.
The evaluation adopts motion-to-motion retrieval metrics, including recall precision at rank $k$ (\ie, \textit{R@1}, \textit{R@2}, \textit{R@3}) and average ranking (\ie, \textit{AvgR}), to assess how accurately the target (\textit{generated-to-target retrieval}) or source motion (\textit{generated-to-source retrieval}) can be retrieved by the generated motion.
We compare our method with baseline methods from \citet{athanasiou2024motionfix}, including MDM, MDM\textsubscript{S}, MDM-BP, MDM-BP\textsubscript{S}, and TMED. The subscript \textsubscript{S} indicates that the model initializes with the source motion instead of random noise for denoising. The ``BD'' label denotes that the model uses a large language model to identify body parts irrelevant to the edit text, keeping these parts fixed via masking.

\paragraph{Results} As shown in \cref{tab:motionfix}, our method outperforms the baselines in both \textit{generated-to-target retrieval} and \textit{generated-to-source retrieval} precision. We emphasize that \textit{generated-to-target retrieval} is the primary performance metric, as it reflects how closely the generated results align with the target. Our representation's inherent multi-scale and compositional properties facilitate learning the relationship between body part dynamics and editing text, significantly improving the accuracy of editing results. %
\cref{fig:motionfix} visually compares TMED~\citep{athanasiou2024motionfix} and our method, showing that our approach captures the edited parts more accurately than TMED, demonstrating its superior performance.

\begin{figure*}[t!]
    \centering
    \includegraphics[width=\linewidth]{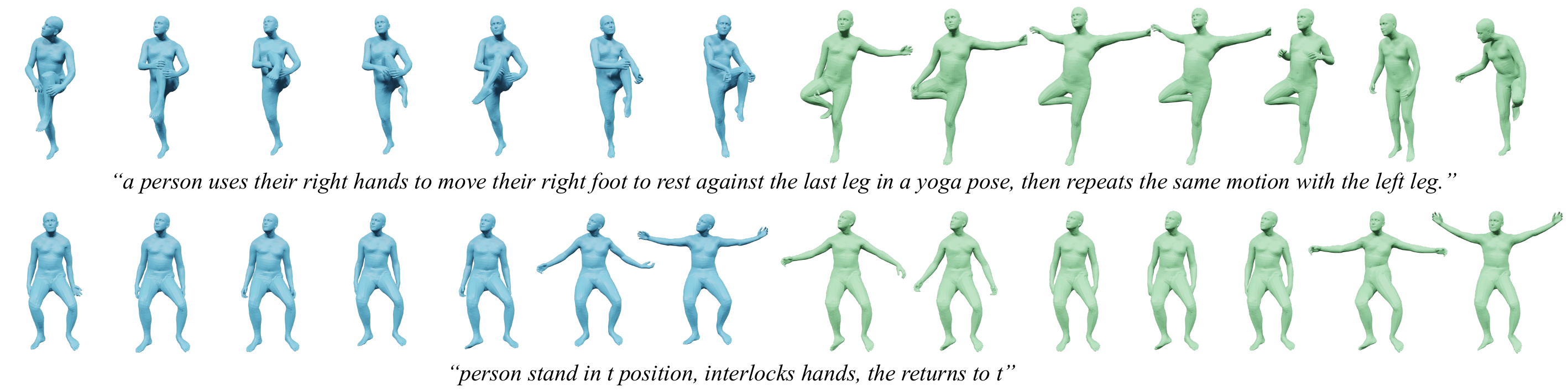}
    \caption{\textbf{Generation results on HumanML3D evaluation set.} \textcolor{motion-blue}{MoMask}~\citep{guo2024momask} fails to generate complex composite motions like ``perform one action first then follow with another'', whereas \textcolor{motion-green}{our results} successfully capture the described dynamics.}
    \label{fig:h3d_generation}
\end{figure*}

\subsection{Evaluation on HumanML3D}
\label{sec:eval_h3d}

\paragraph{Setting, Metrics, and Baselines}
We evaluate our method on one of the most popular text-to-motion benchmarks, \ie, HumanML3D~\citep{guo2022generating}, which contains $14,616$ motion sequences, each described by three text descriptions. In addition to standard metrics such as \textit{FID}, \textit{R-Precision}, \textit{MM Dist.}, \textit{Diversity}, and \textit{MultiModality}, we report the \textit{TMR Similarity Score (TMR)} by following \citet{petrovich2023tmr} to evaluate text-motion similarity using contrastive learning-trained encoders, which are proven more effective in capturing complex text-motion relationships. We also compute a \textit{Success Rate (Succ. Rate)}, considering a test case successful if its TMR similarity score exceeds a threshold of $0.7$.
We report the mean per joint position error (MPJPE) in millimeters when evaluating the reconstruction quality of \model.
The comparison baselines include T2M~\citep{guo2022generating}, MDM~\citep{tevet2023human}, MLD~\citep{chen2023executing}, T2M-GPT~\citep{zhang2023generating}, and MoMask~\citep{guo2024momask}, with which we adopt the same pose representation and dataset split.

\begin{table}[t!]
    \centering
    \small
    \caption{\textbf{Quantitative results of text-to-motion on the HumanML3D~\citep{guo2022generating}.} ``Real'' denotes results computed from ground truth. $\pm$ indicates the $95\%$ confidence interval based on twenty repeated evaluations. \textbf{Bold} indicates the best result. ``$\dag$'' denotes the reproduced results, which are slightly inconsistent with the paper-reported results (shown in \textcolor{gray}{gray} color). The issue is discussed in \href{https://github.com/EricGuo5513/momask-codes/issues/27}{https://github.com/EricGuo5513/momask-codes/issues/27}.}
    \label{tab:h3d}
    \resizebox{\linewidth}{!}{%
        \begin{tabular}{ccccccccc}%
            \toprule
            Model & \makecell{R-Precision\\(Top 1)} $\uparrow$ & FID $\downarrow$ & MM Dist. $\downarrow$ & MultiModality $\uparrow$ & TMR $\uparrow$ & Succ. Rate $\uparrow$ \\
            \midrule
            Real                                            & $0.511^{\pm.003}$                   & $0.002^{\pm.000}$                   & $2.974^{\pm.008}$                   & -                                  & $0.530$ & $0.305$ \\
            \midrule
            T2M~\citep{guo2022generating}                   & $0.457^{\pm.002}$                   & $1.067^{\pm.002}$                   & $3.340^{\pm.008}$                   & $2.090^{\pm.083}$                  & - & -       \\
            T2M-GPT~\citep{zhang2023generating}             & $0.491^{\pm.003}$                   & $0.116^{\pm.004}$                   & $3.118^{\pm.011}$                   & $1.865^{\pm.011}$                  & - & -       \\
            MDM~\citep{tevet2023human}                      & $0.319^{\pm.005}$                   & $0.544^{\pm.044}$                   & $5.566^{\pm.027}$                   & $\mathbf{2.799^{\pm.072}}$         & $0.378^{\pm.005}$                   & $0.079^{\pm.003}$ \\
            MLD~\citep{chen2023executing}                   & $0.481^{\pm.003}$                   & $0.473^{\pm.013}$                   & $3.196^{\pm.010}$                   & $2.413^{\pm.079}$                  & $0.437^{\pm.003}$                   & $0.191^{\pm.004}$ \\
            MoMask$^\dag$~\citep{guo2024momask}             & $\mathbf{0.504^{\pm.003}}$          & $0.124^{\pm.006}$                   & $\mathbf{3.042^{\pm.010}}$          & $1.313^{\pm.044}$                  & $0.503^{\pm.006}$                   & $0.285^{.004}$ \\
            \textcolor{gray}{MoMask~\citep{guo2024momask}}  & \textcolor{gray}{$0.521^{\pm.002}$} & \textcolor{gray}{$0.045^{\pm.002}$} & \textcolor{gray}{$2.958^{\pm.008}$} & \textcolor{gray}{$1.241^{\pm.040}$}& \textcolor{gray}{$0.531^{\pm.005}$} & \textcolor{gray}{$0.289^{\pm.002}$} \\
            \midrule
            Ours                                            & $0.483^{\pm.005}$                   & $\mathbf{0.063^{\pm.007}}$          & $3.094^{\pm.019}$                   & $1.290^{\pm.041}$                  & $\mathbf{0.521^{\pm.003}}$          & $\mathbf{0.297^{\pm.006}}$ \\
            \bottomrule
        \end{tabular}%
    }%
\end{table}

\paragraph{Results}
We present the quantitative results in \cref{tab:h3d}, where our model achieves the best FID, TMR, and Succ. Rate, demonstrating the effectiveness of our proposed methods. The discrepancy between R-Precision and Succ. Rate further suggests significant variance in these metrics when evaluating the consistency between generated motions and text descriptions, highlighting the limitations of these metrics.
The qualitative results on the HumanML3D test set are shown in \cref{fig:h3d_generation}. MoMask~\citep{guo2024momask} struggles to generate motions where the left and right legs move successively or where two consecutive T-poses are performed. In contrast, our method successfully generates plausible dynamics for such complex compositional descriptions.

\subsection{Evaluation on HUMANISE}

\paragraph{Setting, Metrics, and Baselines}
HUMANISE~\citep{wang2022humanise} introduces a more challenging conditional motion generation task, \ie, language-guided human motion generation in 3D scenes, which necessitates a joint modeling of text, 3D scenes, and human motion.
Our evaluation follows the setting from \citet{wang2024move}. We report results using metrics: \textit{goal dist.} for measuring the grounding accuracy of the target interactive object, along with \textit{contact score} and \textit{non-collision score} for evaluating the physical plausibility of the generated motion within 3D scenes.
The baselines include all methods compared in \citet{wang2024move}. Notably, \citet{wang2024move} introduces a two-stage framework that leverages the scene affordance as an intermediary. We replace their second-stage model with our methods as our solution.

\begin{wraptable}[12]{r}{0.65\linewidth}
    \vspace{-2pt}
    \centering
    \small
    \caption{\textbf{Quantitative results of motion generation on the HUMANISE~\citep{wang2022humanise}.} \textbf{Bold} indicates the best result. Each evaluation is repeated five times with a $95\%$ confidence interval indicated by $\pm$.}
    \label{tab:humanise}
    \resizebox{\linewidth}{!}{%
        \begin{tabular}{ccccc}%
            \toprule
            Model &  goal dist. $\downarrow$ & contact $\uparrow$ & non-collision $\uparrow$ \\
            \midrule
            \ac{cvae}~\citep{wang2022humanise} & $0.422^{\pm.011}$          & $84.06^{\pm.716}$          & $99.77^{\pm.004}$          \\
            one-stage@Enc~\citep{wang2024move} & $0.326^{\pm.013}$          & $76.11^{\pm.684}$          & $99.71^{\pm.014}$          \\
            one-stage@Dec~\citep{wang2024move} & $0.185^{\pm.014}$          & $86.43^{\pm.845}$          & $99.76^{\pm.006}$          \\
            two-stage@Enc~\citep{wang2024move} & $0.156^{\pm.006}$          & $95.86^{\pm.323}$          & $99.69^{\pm.007}$          \\
            two-stage@Dec~\citep{wang2024move} & $0.156^{\pm.006}$          & $\mathbf{96.04^{\pm.298}}$ & $99.70^{\pm.005}$          \\
            \midrule
            Ours                               & $\mathbf{0.125^{\pm.004}}$ & $93.10^{\pm.180}$          & $\mathbf{99.78^{\pm.010}}$ \\
            \bottomrule
        \end{tabular}%
    }%
\end{wraptable}
\paragraph{Results}
\cref{tab:humanise} presents the quantitative results. Our method outperforms all baselines regarding $\textit{goal dist.}$ and $\textit{non-collision}$ scores, indicating our representation's effectiveness in the joint modeling of motion, text, and 3D scene. This effective modeling enhances the accuracy of grounding target interactive objects and improves scene awareness for collision-free motion generation.

\section{Conclusion}

This paper targets flexible motion representation by proposing a novel quantization method, which compresses motion sequences into a multi-scale, compositional discrete token space. We highlight the inherent composition capability of this representation by directly composing tokens from different motion sequences. Additionally, we incorporate generative masked modeling to support diverse motion modeling tasks, including motion editing, control, and conditional generation, and demonstrate its superior performance across multiple benchmarks.

\paragraph{Limitations}
Our multi-scale quantization approach results in a larger number of tokens compared to typical frame-level quantization methods like T2M-GPT~\citep{zhang2023generating}, making the second stage of transformer-based inference computationally intensive. Additionally, the manual definition of spatial scales may lead to suboptimal representations.

\paragraph{Acknowledgments} This work is supported in part by the National Science and Technology Major Project (2022ZD0114900) and the National Natural Science Foundation of China (NSFC) (62172043).

\clearpage
{
\small
\bibliographystyle{plainnat}
\bibliography{reference}
}

\clearpage
\appendix
\renewcommand\thefigure{A\arabic{figure}}
\setcounter{figure}{0}
\renewcommand\thetable{A\arabic{table}}
\setcounter{table}{0}
\renewcommand\theequation{A\arabic{equation}}
\setcounter{equation}{0}

\section*{Appendix}

\section{\model Algorithm}

We present the encoding and decoding algorithm of \model in \cref{alg:encoding_decoding}.
Here, $\texttt{Inter}(\bz, n_s)$ denotes the interpolation operation that interpolates 
$\bz \in \bbR^{n \times d}$ to $\barbz_s \in \bbR^{n_s \times d}$, $\texttt{Concat}(\cdot)$ represents the concatenation operation, and $\texttt{FSQ-lookup}(\by_s)$ dequantizes the token indices $\by_s$ into codes $\tilbz_s$.

\begin{algorithm}[ht!]
    \small
    \caption{Encoding and Decoding of \model}
    \label{alg:encoding_decoding}
    
    \texttt{// Encoding Process} \\
    \KwIn{Raw motion sequence $\bx$;}
    Encoders $\left\{\cE_{s}\right\}_{s=1}^{S}$, $f_{0}=\bzero$, $\by=[]$; \\
    Decompose $\bx$ into $\left\{\barbx_s\right\}_{s=1}^{S}$; \\
    \For{$s=1,\cdots,S$}{
        $\bz_s = \cE_s(\barbx_s)$; \\
        $\barbz_s = \texttt{Inter}(\bz_s - f_{s-1}, n_s)$; \\
        $\tilbz_s, \by_s = \texttt{FSQ}(\barbz_s)$; \\
        $\by=\texttt{Concat}(\by, \by_s)$; \texttt{// Concatenation} \\
        $\hatbz_s = \texttt{Inter}(\tilbz_s, n)$; \\
        $f_s=\sum_{i=1}^s\hatbz_i$; \\
    }
    \KwOut{$\by$;}
    \BlankLine
    \texttt{// Decoding Process} \\
    \KwIn{Motion token sequence $\by$;}
    Decoder $\cD$, $\hatf=\bzero$; \\
    Decompose $\by$ into $\left\{\by_s\right\}_{s=1}^S$; \\
    \For{$s=1, \cdots, S$}{
        $\tilbz_s = \texttt{FSQ-lookup}(\by_s)$; \texttt{// Dequantization} \\
        $\hatbz_s = \texttt{Inter}(\tilbz_s, n)$; \\
        $\hatf = \hatf + \hatbz_s$; \\
    }
    $\hatbx = \cD(\hatf)$; \\
    \KwOut{$\hatbx$;}
\end{algorithm}

\section{Implementation Details}
\label{sec:details}

\subsection{\model}

Each scale feature represents a different level of spatial granularity in the motion sequence. Specifically, we decompose the raw motion $\bx$ into $\left\{\barbx_s\right\}_{s=1}^S$ based on body parts by following prior works~\citep{pi2023hierarchical,wan2024tlcontrol}, with each $\barbx_s$ capturing distinct features associated with specific body parts. For all scales, \cref{tab:supp:msq} defines the body parts they encompass.
Besides, each scale's latent feature $\bz_s \in \bbR^{n \times d}$ will be interpolated into $\barbz_s \in \bbR^{n_s \times d}$ along the temporal dimension to construct the multi-scale structure. \cref{tab:supp:msq} also specifies the value of $n_s$ for each scale.
\textbf{Notably, we empirically define the scale for capturing specific body parts and the values of $n_s$ for temporal interpolation. Alternative definitions are feasible but may slightly affect the final performance.}
Each scale possesses a distinct quantizer in our implementation, thus resulting in varying quantization spaces across scales. By default, our implementation uses $6$ scales. \cref{tab:supp:msq} also provides the configurations for the ablative models with varying numbers of scales used in the ablation study presented below. The model with $8$ scales extends the default configuration by repeating the last scale twice.

\begin{table}[t!]
    \centering
    \small
    \caption{\textbf{The detailed definition of each scale in \model.}}
    \label{tab:supp:msq}
        \begin{tabular}{cccccccc}%
            \toprule
            \multirow{2.4}{*}{Model} & \multirow{2.4}{*}{Scale} & \multicolumn{5}{c}{Body Part} & \multirow{2.4}{*}{$n_s$} \\
            \cmidrule(rl){3-7} & & Pelvis & Torso & Legs & Arms & Head & \\
            \midrule
            \multirow{2}{*}{\makecell{Two Scales}}
            & 1 & \cmark &        &        &        &        & 16 \\
            & 2 & \cmark & \cmark & \cmark & \cmark &       \cmark & 49 \\
            \midrule
            \multirow{4}{*}{\makecell{Four Scales}}
            & 1 & \cmark & \cmark &        &        &        & 16 \\
            & 2 & \cmark & \cmark & \cmark &        &        & 24 \\
            & 3 & \cmark & \cmark & \cmark & \cmark &        & 32 \\
            & 4 & \cmark & \cmark & \cmark & \cmark &       \cmark & 49 \\
            \midrule
            \multirow{6}{*}{\makecell{Six Scales\\(default)}}
            & 1 & \cmark &        &        &        &        & 16 \\
            & 2 & \cmark & \cmark &        &        &        & 24 \\
            & 3 & \cmark & \cmark & \cmark &        &        & 32 \\
            & 4 & \cmark & \cmark & \cmark & \cmark &        & 40 \\
            & 5 & \cmark & \cmark & \cmark & \cmark & \cmark & 49 \\
            & 6 & \cmark & \cmark & \cmark & \cmark & \cmark & 49 \\
            \bottomrule
        \end{tabular}%
\end{table}

\subsection{Masked Transformer}

Once the motion $x$ is quantized into discrete index $\by=\{\by_s\}_{s=1}^{S}$, we flatten $\by$ to form a token index sequence. For simplicity, we reuse $\by$ to denote this flattened index sequence and assume its length is $n$. During the training of the masked transformer, we first randomly mask $\lceil \gamma(\tau) \cdot n \rceil$ tokens according to the masking schedule function $\gamma(\tau)=\cos(\frac{\pi \tau}{2})$, and replace them with \texttt{[MASK]} token. Here, $\tau \in [0, 1)$ is uniformly sampled from $0$ to $1$, and $\tau=0$ means all takens are masked. Then, the masked transformer learns to reconstruct the masked tokens by minimizing the following negative log-likelihood:
\begin{equation}
\mathcal{L} = \sum_{\tilde{\by}_i = \texttt{[MASK]}} -\log p_{\theta} (\by_i | \tilde{\by}, \mathcal{C})
\end{equation}
Here, $\tilde{\by}$ is the sequence after masking, $i$ is the entry index, $\theta$ is model parameters, and $\mathcal{C}$ is the input condition (\eg, text). During the inference, the masked transformer starts with a fully masked token sequence and generates all tokens auto-regressively, as described in the main paper.

The model architecture builds on the implementation by \citet{zhang2023generating}, while the masked modeling framework adopts the code from \citet{guo2024momask}. \textbf{We will release our code for reproducibility.}

\begin{table*}[t!]
    \centering
    \small
    \caption{\textbf{Quantitative results of text-to-motion on the HumanML3D~\citep{guo2022generating}.} ``Real'' denotes results computed from ground truth. ``$\rightarrow$'' indicates metrics are better when closer to the ``Real''. $\pm$ indicates the $95\%$ confidence interval based on twenty repeated evaluations.}
    \label{supp:tab:h3d}
    \resizebox{\linewidth}{!}{%
        \begin{tabular}{cccccccccc}%
            \toprule
            \multirow{2.4}{*}{Model} & \multicolumn{3}{c}{R-Precision $\uparrow$} & \multirow{2.4}{*}{FID $\downarrow$} & \multirow{2.4}{*}{MM Dist. $\downarrow$} & \multirow{2.4}{*}{Diversity $\rightarrow$} & \multirow{2.4}{*}{MultiModality $\uparrow$} & \multirow{2.4}{*}{TMR $\uparrow$} & \multirow{2.4}{*}{Succ. Rate $\uparrow$}\\
            \cmidrule(rl){2-4} & Top 1 & Top 2 & Top 3 & \\
            \midrule
            Real                                            & $0.511^{\pm.003}$          & $0.703^{\pm.003}$          & $0.797^{\pm.002}$          & $0.002^{\pm.000}$          & $2.974^{\pm.008}$          & $9.503^{\pm.065}$          & -                          & $0.530$ & $0.305$ \\
            \midrule
            T2M~\citep{guo2022generating}                   & $0.457^{\pm.002}$          & $0.639^{\pm.003}$          & $0.740^{\pm.003}$          & $1.067^{\pm.002}$          & $3.340^{\pm.008}$          & $9.188^{\pm.002}$          & $2.090^{\pm.083}$          & - & - \\
            T2M-GPT~\citep{zhang2023generating}             & $0.491^{\pm.003}$          & $0.680^{\pm.003}$          & $0.775^{\pm.002}$          & $0.116^{\pm.004}$          & $3.118^{\pm.011}$          & $9.761^{\pm.081}$          & $1.865^{\pm.011}$          & - & - \\
            MDM~\citep{tevet2023human}                      & $0.319^{\pm.005}$          & $0.498^{\pm.004}$          & $0.611^{\pm.007}$          & $0.544^{\pm.044}$          & $5.566^{\pm.027}$          & $9.559^{\pm.086}$          & $2.799^{\pm.072}$          & $0.378^{\pm.005}$ & $0.079^{\pm.003}$ \\
            MLD~\citep{chen2023executing}                   & $0.481^{\pm.003}$          & $0.673^{\pm.003}$          & $0.772^{\pm.002}$          & $0.473^{\pm.013}$          & $3.196^{\pm.010}$          & $9.724^{\pm.082}$          & $2.413^{\pm.079}$          & $0.437^{\pm.003}$ & $0.191^{\pm.004}$ \\
            MoMask (Base)$^\dag$~\citep{guo2024momask}
            & $0.503^{\pm.002}$                   & $0.693^{\pm.002}$                   & $0.791^{\pm.002}$                   & $0.137^{\pm.006}$                   & $3.056^{\pm.006}$                   & $9.689^{\pm.089}$                   & $1.102^{\pm.040}$                   & $0.529^{\pm.004}$ & $0.289^{\pm.003}$ \\
            \textcolor{gray}{MoMask (Base)~\citep{guo2024momask}}
            & \textcolor{gray}{$0.504^{\pm.004}$} & \textcolor{gray}{$0.699^{\pm.006}$} & \textcolor{gray}{$0.797^{\pm.004}$} & \textcolor{gray}{$0.082^{\pm.008}$} & \textcolor{gray}{$3.050^{\pm.013}$} & -                                   & \textcolor{gray}{$1.050^{\pm.061}$} & \textcolor{gray}{$0.527^{\pm.003}$} & \textcolor{gray}{$0.283^{\pm.002}$}\\
            MoMask (Full)$^\dag$~\citep{guo2024momask}
            & $0.504^{\pm.003}$                   & $0.697^{\pm.002}$                   & $0.795^{\pm.003}$                   & $0.124^{\pm.006}$                   & $3.042^{\pm.010}$                   & $9.302^{\pm.077}$                   & $1.313^{\pm.044}$                   & $0.503^{\pm.006}$ & $0.285^{\pm.004}$ \\
            \textcolor{gray}{MoMask (Full)~\citep{guo2024momask}}
            & \textcolor{gray}{$0.521^{\pm.002}$} & \textcolor{gray}{$0.713^{\pm.003}$} & \textcolor{gray}{$0.807^{\pm.002}$} & \textcolor{gray}{$0.045^{\pm.002}$} & \textcolor{gray}{$2.958^{\pm.008}$} & \textcolor{gray}{$9.624^{\pm.080}$} & \textcolor{gray}{$1.241^{\pm.040}$} & \textcolor{gray}{$0.531^{\pm.005}$} & \textcolor{gray}{$0.289^{\pm.002}$} \\
            \midrule
            Ours                                            & $0.483^{\pm.005}$          & $0.683^{\pm.002}$          & $0.778^{\pm.002}$          & $0.063^{\pm.007}$          & $3.094^{\pm.019}$          & $9.497^{\pm.204}$          & $1.290^{\pm.041}$          & $0.521^{\pm.003}$ & $0.297^{\pm.006}$ \\
            \bottomrule
        \end{tabular}%
    }%
\end{table*}

\subsection{Evaluation Details}

For the evaluation on the MotionFix, HumanML3D, and HUMANISE benchmarks, we train our model from scratch using the corresponding benchmark-provided data, without any pretraining on other datasets. Below, we outline specific evaluation details for each benchmark.

\paragraph{Evaluation Details on MotionFix}
To train our masked transformer on MotionFix, we use the source motion token sequence as input, allowing the model to recover the target token sequence in a single iteration (\ie, $K=1$). Optionally, partial source motion tokens corresponding to the edited regions can be masked by leveraging ChatGPT to parse the editing text description. In our implementation, we use the original source motion tokens, which have already achieved promising results. 
Since MotionFix has not provided implementation details on using HumanML3D as a training data source, we did not evaluate our model leveraging HumanML3D data for training.

\paragraph{Evaluation Details on HumanML3D}
The standard evaluation metrics include (1) \textit{FID}, measuring the distributional discrepancy between generated and real motions; (2) \textit{R-Precision} and \textit{MM Dist.}, assessing the relevance between the generated results and descriptions; (3) \textit{Diversity}, evaluating the overall variation of all results; (4) \textit{MultiModality}, evaluating the average variation of motions generated from one text. To compute these metrics, we follow prior works by sampling a single description for each test case (each case in HumanML3D has three descriptions) during inference.
Additionally, we introduce the TMR success rate to assess the correspondence between text and generated motion, but using all text descriptions for each test case during inference. Before computing the TMR score, we exclude low-quality descriptions where the TMR score computed with the ground-truth motion is below $0.3$ to ensure evaluation quality. We use CLIP~\citep{radford2021learning} to encode the text.

\paragraph{Evaluation Details on HUMANISE}

For evaluation on HUMANISE, we follow the setting in \citet{wang2024move}, and our method adapts their two-stage model for language-guided human motion generation in 3D scenes. Specifically, \citet{wang2024move} proposes a two-stage framework that leverages scene affordance as an intermediary. We retain the first stage for affordance generation and replace the second-stage model with our masked transformer. We use the same encoder to process the affordance maps and the text descriptions, concatenate the resulting conditional tokens with the masked motion tokens, and pass the concatenation to the masked transformer, which only employs the self-attention mechanism in the design.

\section{Additional Results}

The following text presents additional results of our method. \textbf{We recommend referring to the supplementary video for dynamic motion animations.}

\subsection{Full Quantitative Results on HumanML3D}

\cref{supp:tab:h3d} presents the full quantitative result on HumanML3D.
This table also reports the ``MoMask (base)'' model, which removes the residual transformer. ``MoMask (full)'' model includes these residual transformers for generating residual motion tokens. $\dag$ denotes the reproduced results, and the results in \textcolor{gray}{gray} are original paper reported results.

\begin{figure*}[t!]
    \centering
    \begin{subfigure}{0.33\textwidth}
        \centering
        \includegraphics[width=\linewidth]{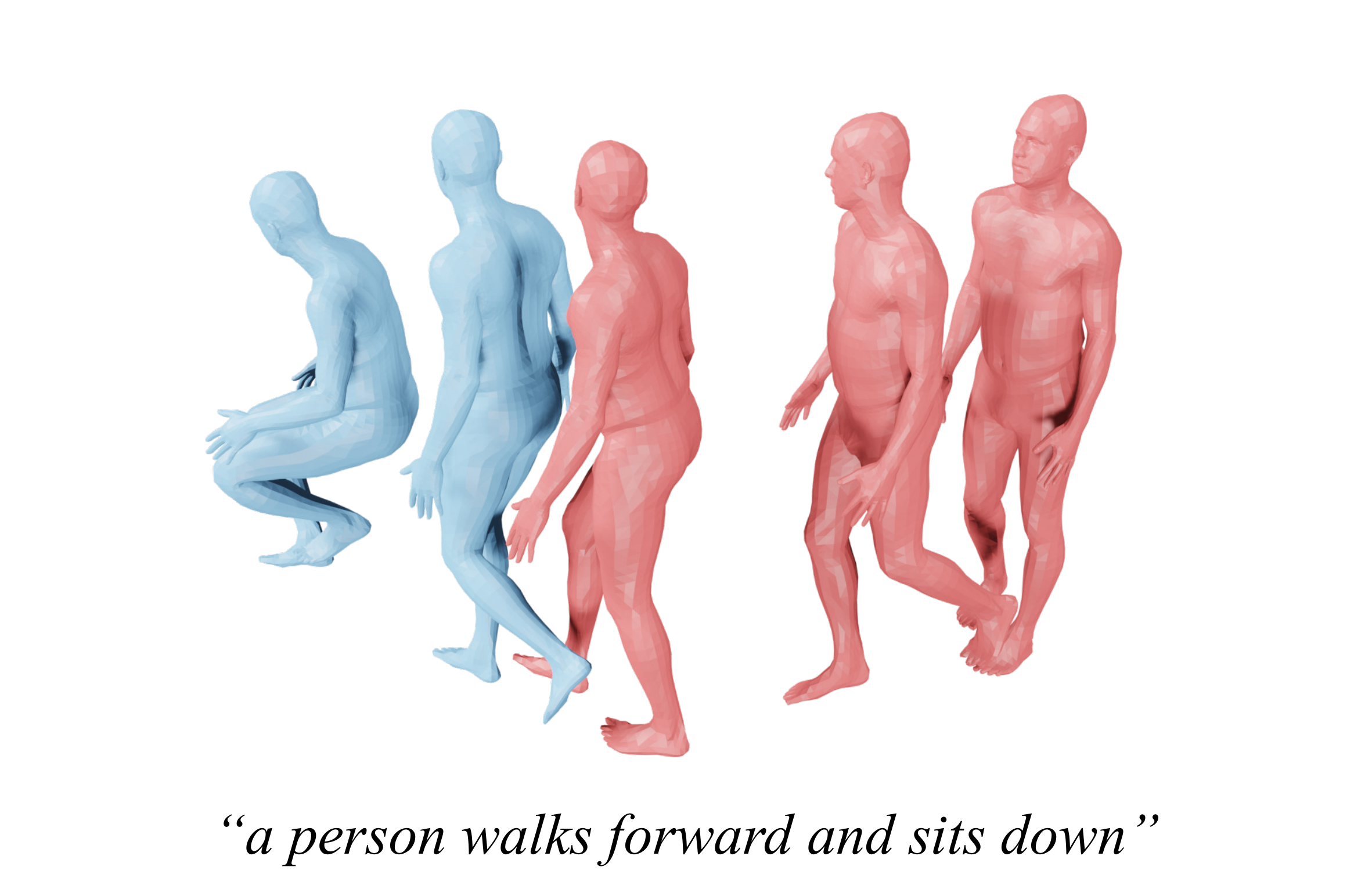}
        \caption{Motion suffix inpainting}
        \label{fig:subfig1}
    \end{subfigure}%
    \begin{subfigure}{0.33\textwidth}
        \centering
        \includegraphics[width=\linewidth]{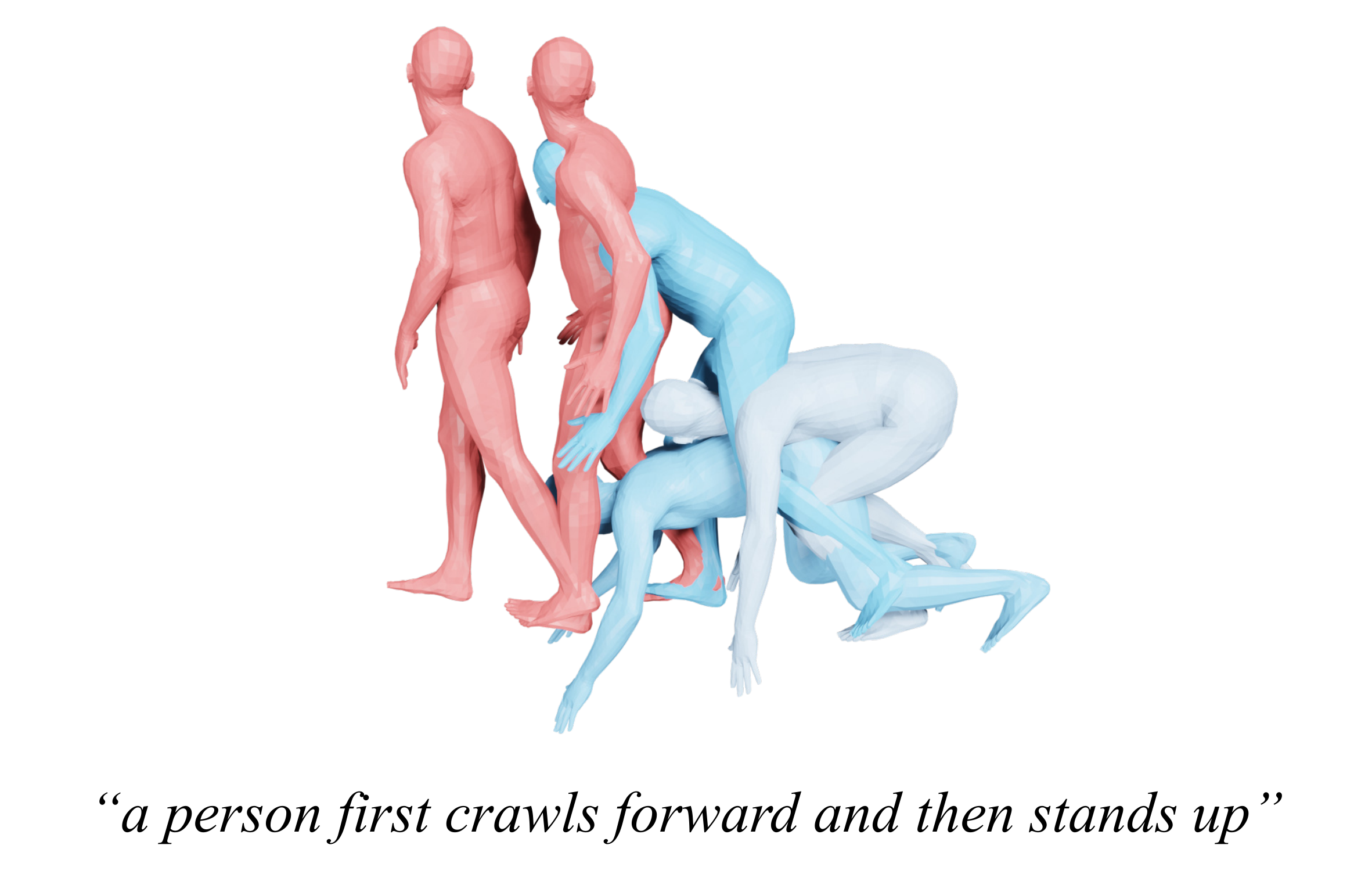}
        \caption{Motion prefix inpainting}
        \label{fig:subfig2}
    \end{subfigure}%
    \begin{subfigure}{0.33\textwidth}
        \centering
        \includegraphics[width=\linewidth]{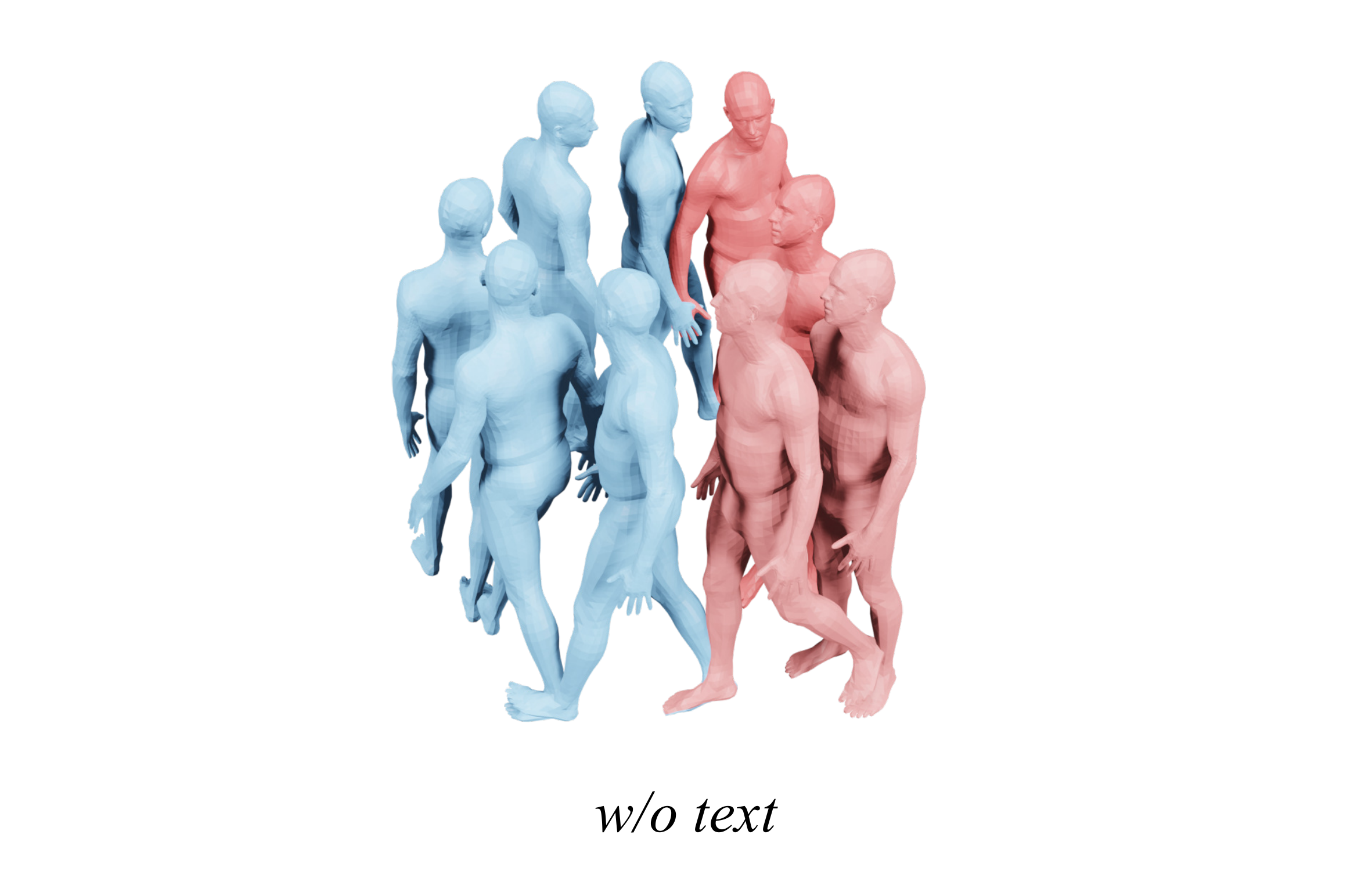}
        \caption{Motion in-between inpainting}
        \label{fig:subfig3}
    \end{subfigure}%
    \caption{\textbf{Temporal motion inpainting.} Poses in \textcolor{motion-red}{red} denotes the given frames, and poses in \textcolor{motion-blue}{bule} are recovered frames.}
    \label{fig:temporal_inpainting}
\end{figure*}

\begin{figure*}[t!]
    \centering
    \begin{subfigure}{0.5\textwidth}
        \centering
        \includegraphics[width=\linewidth]{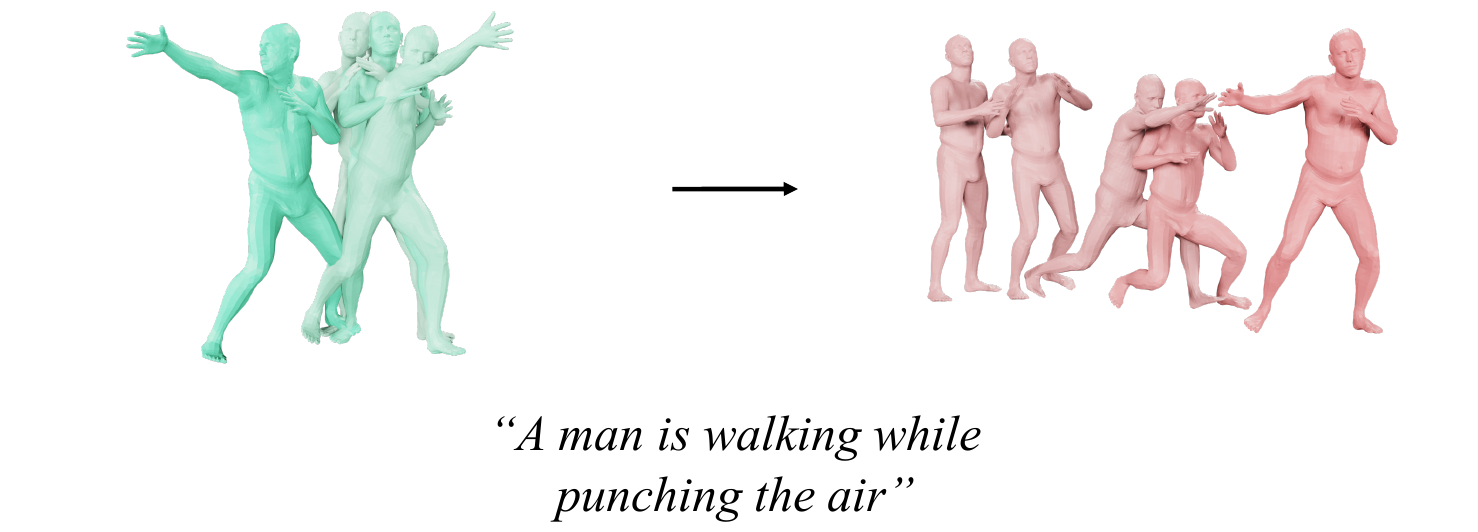}
        \caption{Lower body inpainting}
        \label{fig:spatial_subfig1}
    \end{subfigure}%
    \begin{subfigure}{0.5\textwidth}
        \centering
        \includegraphics[width=\linewidth]{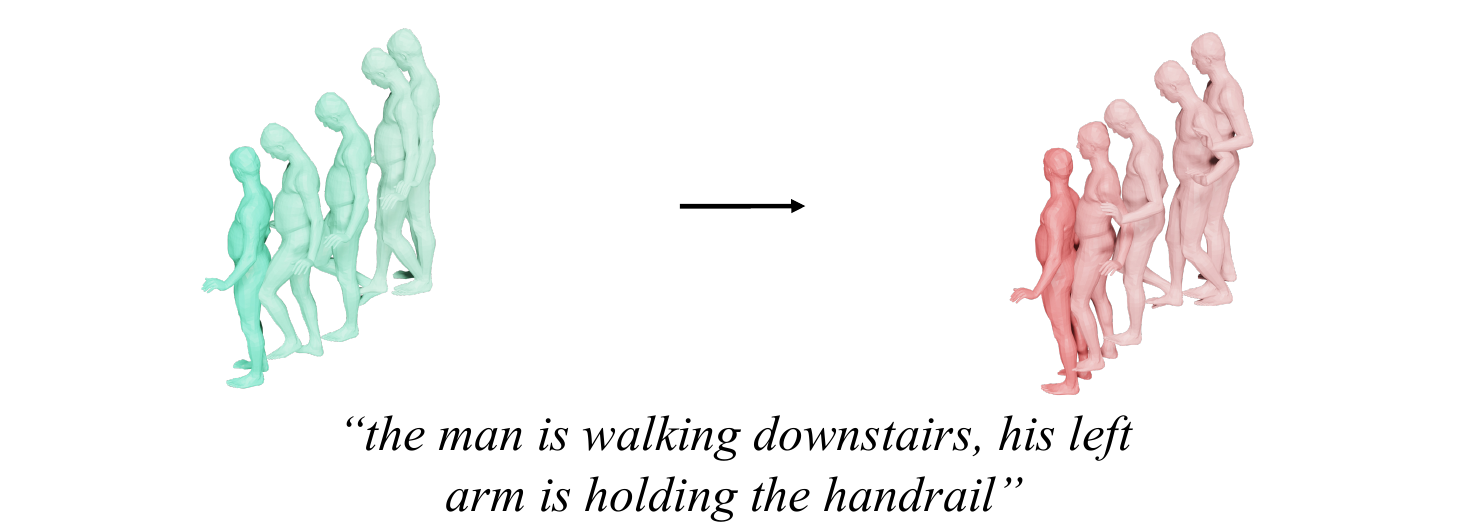}
        \caption{Upper body inpainting}
        \label{fig:spatial_subfig2}
    \end{subfigure}%
    \caption{\textbf{Spatial motion inpainting.} Motions in \textcolor{motion-green}{green} are the source motions, and motions in \textcolor{motion-red}{red} are the inpainting results.}
    \label{fig:spatial_inpainting}
\end{figure*}

\begin{table}[t!]
    \centering
    \small
    \setlength{\tabcolsep}{3pt}
    \caption{\textbf{Quantitative results on five motion inpainting tasks.}}
    \label{tab:inpainting}
        \begin{tabular}{cccccc}%
            \toprule
            Task & \makecell{R-Precision\\(Top 1)} $\uparrow$ & FID $\downarrow$ & MM Dist. $\downarrow$ & Diversity $\rightarrow$ \\
            \midrule
            In-betweening       & $0.496^{\pm.003}$ & $0.067^{\pm.002}$ & $3.004^{\pm.007}$ & $9.331^{\pm.087}$ \\
            Suffix Inpainting   & $0.498^{\pm.003}$ & $0.107^{\pm.002}$ & $3.004^{\pm.007}$ & $9.264^{\pm.066}$ \\
            Prefix Inpainting   & $0.506^{\pm.003}$ & $0.062^{\pm.002}$ & $2.978^{\pm.008}$ & $9.416^{\pm.084}$ \\
            \midrule
            Upper Body Editing  & $0.499^{\pm.004}$ & $0.115^{\pm.003}$ & $3.032^{\pm.009}$ & $9.432^{\pm.097}$ \\
            Lower Body Editing  & $0.473^{\pm.003}$ & $0.210^{\pm.004}$ & $3.138^{\pm.007}$ & $9.219^{\pm.077}$ \\
            \bottomrule
        \end{tabular}%
\end{table}

\subsection{Motion Inpainting Results}

Our method supports both temporal and spatial inpainting of the motion sequence, and we evaluate the inpainting capability of our method across five tasks: motion in-betweening, suffix inpainting, prefix inpainting, upper body inpainting, and lower body inpainting. Specifically, during testing, we replace the partial frames or body part features with zero vectors, quantize the incomplete sequences using \model, and apply the pre-trained bidirectional masked transformer for recovery.
For temporal inpainting, we mask $70\%$ of the frames in the middle of the motion sequence for motion in-betweening, $70\%$ of the frames at the beginning for motion prefix inpainting, and $70\%$ of the frames at the end for suffix inpainting. For spatial inpainting, we mask the joint features of the upper or lower body for upper and lower body inpainting tasks.
The motion inpainting results on HumanML3D~\citep{guo2022generating}, presented in \cref{fig:temporal_inpainting}, \cref{fig:spatial_inpainting}, and \cref{tab:inpainting}, highlight the effectiveness of our method in recovering incomplete motions.

\begin{figure*}[t!]
    \centering
    \includegraphics[width=\linewidth]{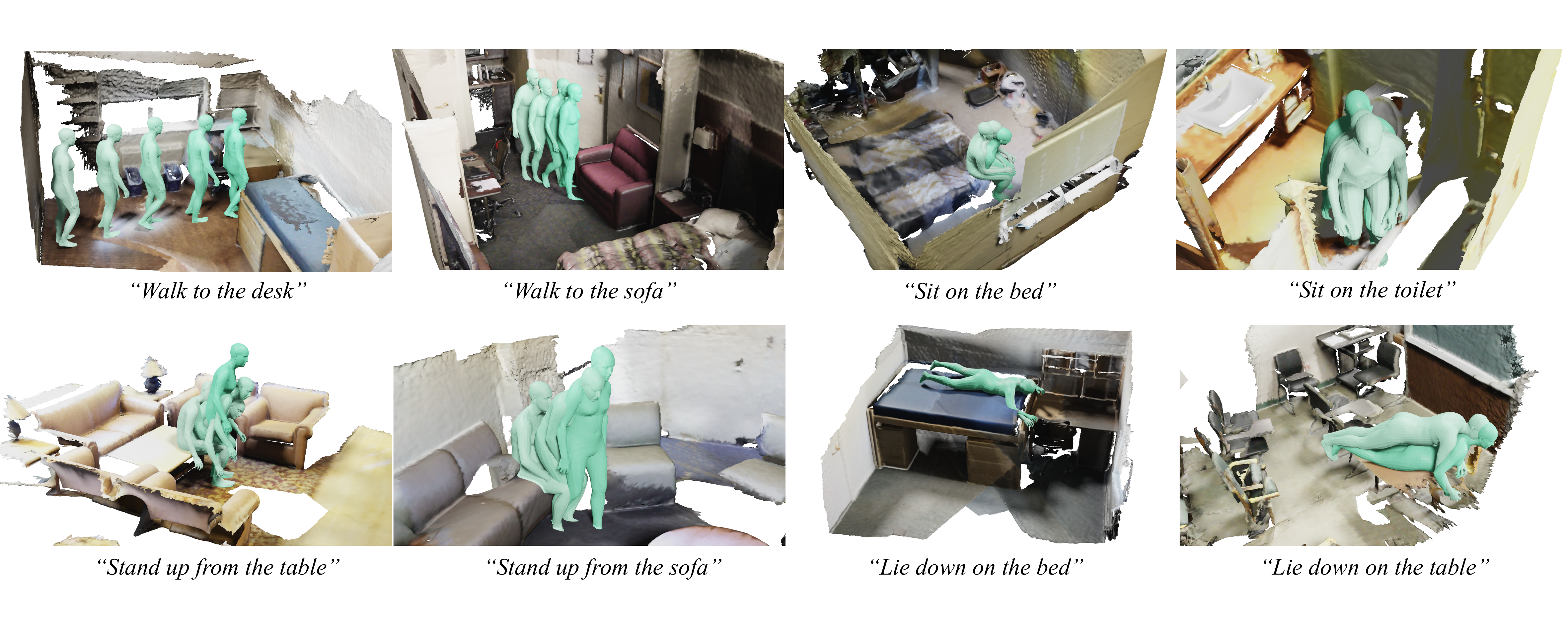}
    \caption{\textbf{Visualization results on HUMANISE.}}
    \label{fig:humanise}
\end{figure*}

\begin{figure}[t!]
    \centering
    \includegraphics[width=0.6\linewidth]{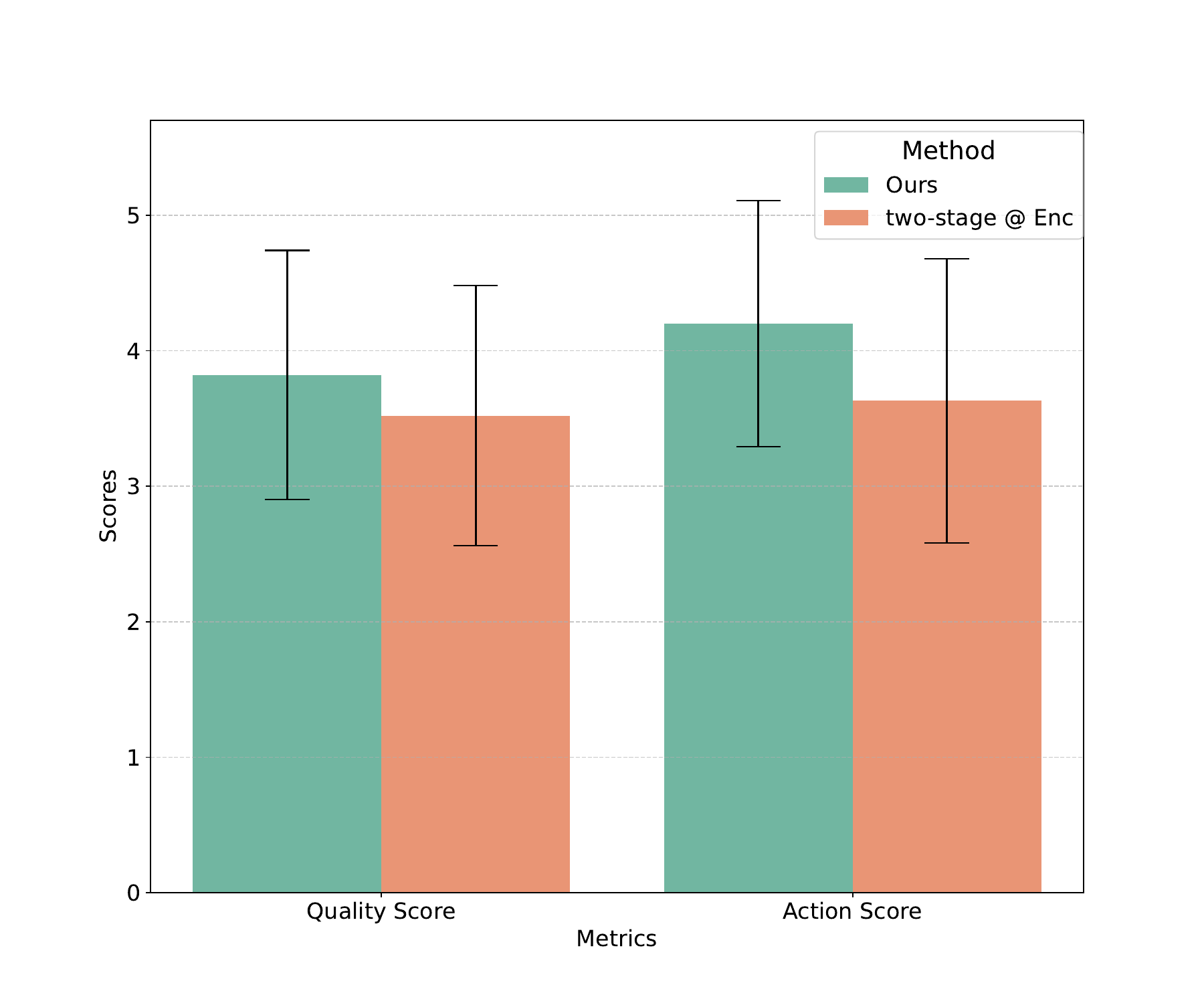}
    \caption{\textbf{User study on HUMANISE.}}
    \label{fig:user_study}
\end{figure}

\subsection{Additional Results on HUMANISE}

\paragraph{Visualization Results} We present the visualization results of our method on the HUMANISE dataset in \cref{fig:humanise}. These results demonstrate that our method effectively grounds target objects and generates physically plausible motion in 3D scenes guided by language descriptions.

\paragraph{User Study} We further evaluate the generation quality through user study. Following HUMANISE~\citep{wang2022humanise}, we ask users to rate the generated motions on a scale from $1$ to $5$ based on (i) overall generation quality (\ie, \textit{quality score}) and (ii) whether the motions accurately perform the action described by the text (\ie, \textit{action score}). Specifically, we generate $60$ human-scene interaction cases using the previous model, ``two-stage @ Enc'' proposed by ~\citet{wang2024move}, and another $60$ cases using our method. These $120$ cases are randomly shuffled and presented to four users. We report the results in \cref{fig:user_study}, indicating that our method outperforms the previous approach in both \textit{quality score} and \textit{action score}.

\subsection{Ablation Studies}

\begin{table}[t!]
    \centering
    \small
    \setlength{\tabcolsep}{3pt}
    \caption{\textbf{Ablation of scale number $S$.}}
    \label{tab:ablation_num_scale}
    \resizebox{0.6\linewidth}{!}{%
        \begin{tabular}{ccccc}%
            \toprule
            Model & \makecell{R-Precision\\(Top 1)} $\uparrow$ & FID $\downarrow$ & MM Dist. $\downarrow$ & MPJPE $\downarrow$ \\
            \midrule
            Two Scales   & $0.501^{\pm.004}$ & $0.109^{\pm.002}$ & $2.980^{\pm.014}$ & $37.68^{\pm.147}$ \\
            Four Scales  & $0.509^{\pm.005}$ & $0.062^{\pm.001}$ & $2.925^{\pm.001}$ & $26.49^{\pm.055}$ \\
            Six Scales   & $0.511^{\pm.000}$ & $0.037^{\pm.000}$ & $2.934^{\pm.012}$ & $23.67^{\pm.050}$ \\
            Eight Scales & $0.513^{\pm.004}$ & $0.018^{\pm.000}$ & $2.912^{\pm.012}$ & $16.41^{\pm.030}$ \\
            \bottomrule
        \end{tabular}%
    }%
\end{table}

We conduct ablation studies to investigate the impact of the scale number $S$. \cref{tab:ablation_num_scale} presents the quantitative results. As $S$ increases, \model demonstrates improved reconstruction quality, as metrics like \textit{FID} and \text{MPJPE} indicate. However, \textit{R-Precision} and \textit{MM Dist.} begin to saturate when the scale number reaches $4$, primarily because these metrics evaluate the alignment between motion and text in the high-level latent space, neglecting fine-grained details.

\section{Licenses for Existing Assets}

We conduct our experiments on the MotionFix~\citep{athanasiou2024motionfix}, HumanML3D~\citep{guo2022generating}, and HUMANISE~\citep{wang2022humanise} datasets. MotionFix is released under the terms of the \href{https://creativecommons.org/licenses/by/4.0/}{Creative Commons Attribution 4.0 International License}, while both HumanML3D and HUMANISE are licensed under the \href{https://opensource.org/license/mit}{MIT License}.

\section{Broader Impacts}

This work on human motion generation holds the potential to positively impact a range of domains, including virtual reality (VR), assistive robotics, animation, and human-computer interaction. 
For instance, more realistic and controllable motion generation can enable more immersive and responsive virtual agents in education, training, or mental health applications.
However, if misused, motion generation technologies may contribute to unethical surveillance systems or the creation of synthetic media that impersonate individuals without their consent.

\end{document}